\documentclass[lettersize,journal]{IEEEtran}
\usepackage{amsmath,amsfonts}
\usepackage{algorithmic}
\usepackage{algorithm}
\usepackage{array}
\usepackage[caption=false,font=normalsize,labelfont=sf,textfont=sf]{subfig}
\usepackage{textcomp}
\usepackage{stfloats}
\usepackage{url}
\usepackage{verbatim}
\usepackage{graphicx}
\usepackage{cite}
\usepackage{color}
\usepackage{xcolor}

\newcommand{\ie}{\textit{i.e.}}
\newcommand{\eg}{\textit{e.g.}}

\usepackage{bbm} 
\usepackage{multirow} 
\usepackage{bbding} 
\usepackage{hyperref} 
\usepackage{float}

\begin{document}

\title{AO-DETR: Anti-Overlapping DETR for X-Ray Prohibited Items Detection}
\author{Mingyuan Li, Tong Jia, Hao Wang, Bowen Ma, Shuyang Lin, Da Cai, and Dongyue Chen
\thanks{This work is supported by the National Natural Science Foundation of China under Grant U22A2063, 62173083, and 62206043; the National Key Research and Development Project of China under No.2022YFF0902401; Guangdong Basic and Applied Basic Research Foundation (2021B1515120064); the Major Program of National Natural Science Foundation of China (71790614) and the 111 Project (B16009). \textit{(Corresponding author: Tong Jia.)}}
\thanks{
Mingyuan Li, Hao Wang, Bowen Ma, Shuyang Lin, Da Cai, and are with the
College of Information Science and Engineering, Northeastern University, Shenyang, 110819, Liaoning, China (e-mail: 542027743@qq.com; ddsywh@yeah.net; 2010285@stu.neu.edu.cn; 2210329@stu.neu.edu.cn; 1553317260@qq.com).}
\thanks{Dongyue Chen is with the College of Information Science and Engineering, Northeastern University, Shenyang, 110819, Liaoning, China, and also with the Foshan Graduate School of Innovation, Northeastern University, Foshan, 528311, Guangdong, China (e-mail: chendongyue@ise.neu.edu.cn).}
\thanks{Tong Jia is with the College of Information Science and Engineering, Northeastern University, Shenyang, 110819, Liaoning, China, and also with the Key Laboratory of Data Analytics and Optimization for Smart Industry, Ministry of Education, Northeastern University, Shenyang, 110819, Liaoning, China (e-mail: jiatong@ise.neu.edu.cn).}
}

   

\maketitle

\begin{abstract}
Prohibited item detection in X-ray images is one of the most essential and highly effective methods widely employed in various security inspection scenarios.
Considering the significant overlapping phenomenon in X-ray prohibited item images, we propose an Anti-Overlapping DETR (AO-DETR) based on one of the state-of-the-art general object detectors, DINO.
Specifically, to address the feature coupling issue caused by overlapping phenomena, we introduce the Category-Specific One-to-One Assignment (CSA) strategy to constrain category-specific object queries in predicting prohibited items of fixed categories, which can enhance their ability to extract features specific to prohibited items of a particular category from the overlapping foreground-background features. 
To address the edge blurring problem caused by overlapping phenomena, we propose the Look Forward Densely (LFD) scheme, which improves the localization accuracy of reference boxes in mid-to-high-level decoder layers and enhances the ability to locate blurry edges of the final layer.
Similar to DINO, our AO-DETR provides two different versions with distinct backbones, tailored to meet diverse application requirements.
Extensive experiments on the PIXray and OPIXray datasets demonstrate that the proposed method surpasses the state-of-the-art object detectors, indicating its potential applications in the field of prohibited item detection.
The source code will be released at \href{https://github.com/Limingyuan001/AO-DETR-test}{https://github.com/Limingyuan001/AO-DETR}.

\end{abstract}


\section{Introduction}
\begin{figure}
    \centering
    \includegraphics[width=0.9\linewidth]{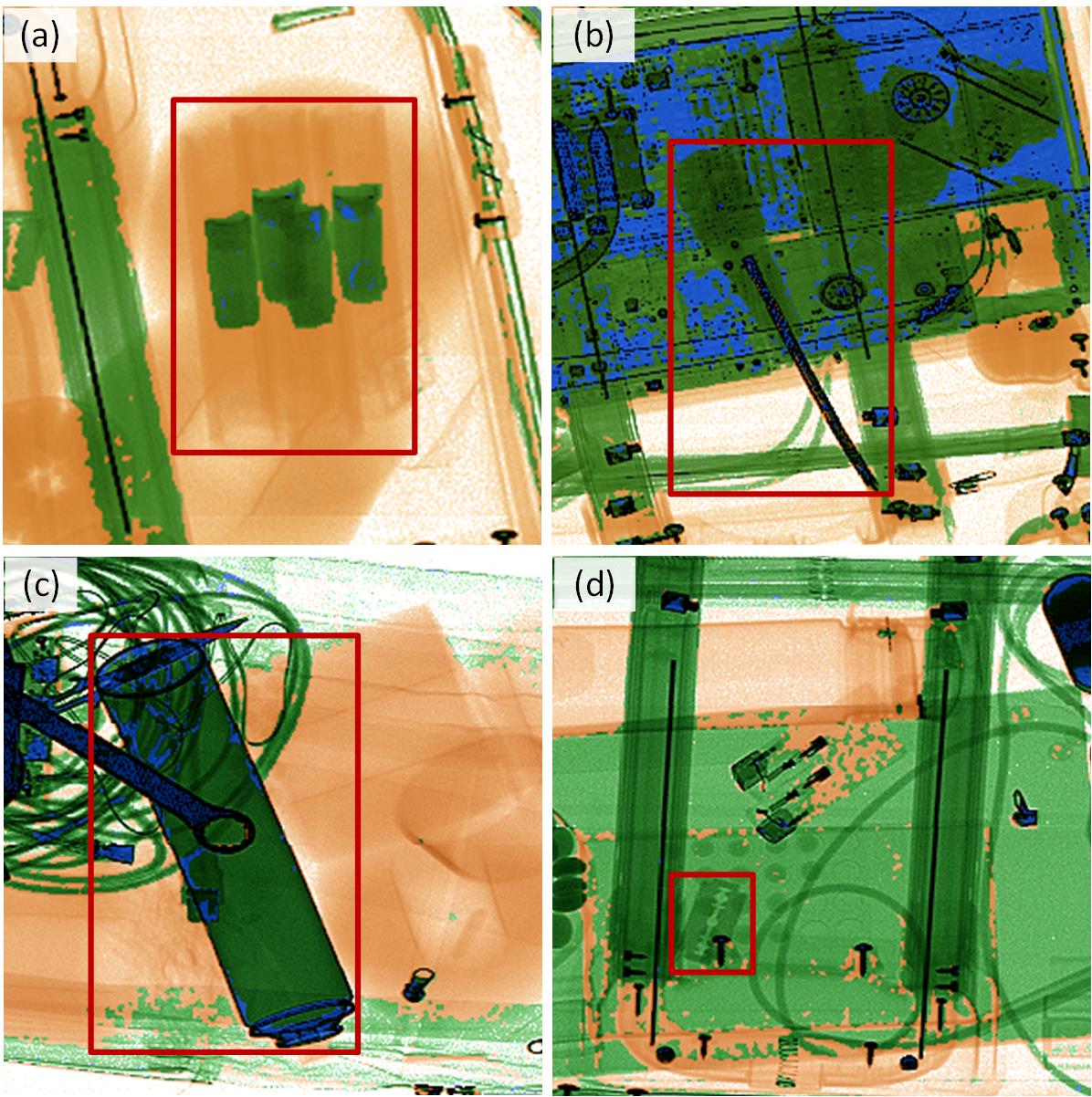}
    \caption{The localized X-ray images with prohibited items. The phenomenon of overlap in images, to varying extents, leads to the overlapping of foreground and background as well as the blurring of object boundaries.
    }
    \label{X-ray images}
\end{figure}

\IEEEPARstart{S}{ECURITY} inspection is one of the most vital and crucial measures to uncovering the potential risks in public spaces, such as airports, train stations, subway stations, and sensitive departments. 
Currently, a predominant approach to contraband detection involves the acquisition of X-ray images of luggage through security scanning machine, followed by a meticulous manual inspection conducted by security staff who have undergone specialized training.
With the advancement of computer vision technology, researchers~\cite{SIXray,OPIXray,PIXray,Xdet,PIDray,tao2022few,shao2022exploiting,PIXDet,PID-YOLOX} have attempted to apply models from the general field of image classification, object detection, and semantic segmentation to the realm of X-ray prohibited item detection, aiming to assist security staff in auxiliary inspections. However, as shown in Fig.~\ref{X-ray images}, X-ray images exhibit the overlapping phenomenon, leading to the coupling of foreground and background information and issues of edge blurring. This overlapping phenomenon reduces the detection accuracy of general detectors.

To enhance the performance of object detectors in X-ray images, two mainstream approaches have been generally utilized, accurate label assignment and multi-stage regression. 
In the accurate label assignment approach, there are several studies, such as GADet~\cite{GADet}, CLCXray~\cite{CLCXray}, and Xdet~\cite{Xdet}, which have introduced precise label assignment methods like IAA, LAreg, and HSS. These methods aim to enhance the capacity of networks to discern features of foreground objects amidst overlapping foreground and background. 
However, they all endeavor to directly augment the capacity of the model to discern features of all foreground categories amidst overlapping features, which is evidently too optimistic, given that the nature and pattern of coupling between each category and background differs.
In the multi-stage regression approach, in an effort to refine edge localization accuracy, numerous models have adopted a sophisticated multi-stage regression framework. Notable examples of such models include the likes of Faster R-CNN~\cite{Faster}, Cascade R-CNN~\cite{Cascade}, and Libra R-CNN~\cite{Libra}.
However, some concepts of them are not fully applicable to the iterative refinements boxes scheme in more advanced detector architectures, \ie, Detection Transformer (DETR)~\cite{DETR}, Deformable DETR~\cite{Deformable-DETR}, and DINO~\cite{DINO}.

In this paper, we try to improve the anti-overlapping capability of the detector without changing the architecture of model DINO~\cite{DINO} and without adding any computational complexity.
In addition, to pursue the optimal performance in X-ray image prohibited item detection field, we introduce a novel object detection model named Anti-Overlapping DETR (AO-DETR), based on DINO, which is the SOTA DETR-like model in natural image object detection field. 
\begin{figure*}
    \centering
    \includegraphics[width=0.8\linewidth]{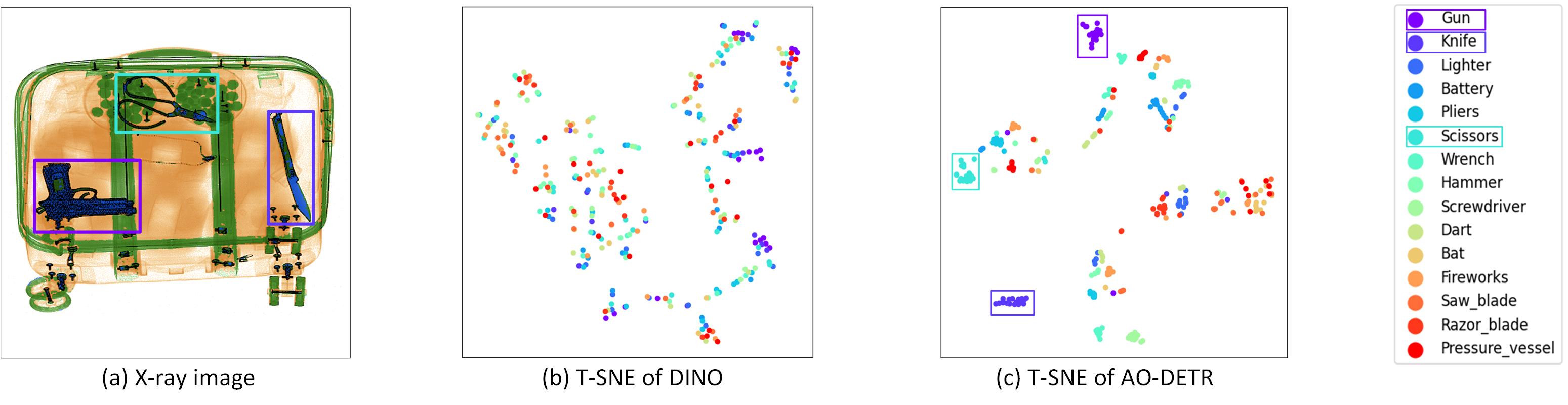}
    \caption{T-SNE dimensionality reduction comparison. (a) The original X-ray image containing a gun, scissors, and a knife. (b)(c) The distributions visualization of T-SNE dimensionality reduction of the object queries from the last decoder layer in DINO and AO-DETR.
    }
    \label{figure T-SNE}
\end{figure*}
Specifically, to alleviate the adverse effects caused by the ambiguity of category semantics in object queries due to the overlapping of foreground and background in X-ray images, as illustrated in Fig~\ref{figure T-SNE}(b), we propose a category-specific one-to-one assignment strategy (CSA). 
Through CSA, the category-specific object queries will be stably assigned to ground truth labels and reference boxes of the specific category of prohibited items, enabling it to specialize in extracting feature of specific category of prohibited items from overlapping foreground and background. 
In addition, to mitigate the blurred boundary problem caused by overlapping phenomena, after conducting a detailed analysis of why look forward twice (LFT) is more suitable for precise boundary localization than Look Forward Once (LFO), we propose the Look Forward Densely (LFD) scheme, which can localize the edges of foreground objects more accurately. 
LFD is capable of transmitting gradients densely to multiple decoder layers, allowing low-level decoder layers to provide more reliable reference boxes to deformable attention in high-level decoder layers. This helps high-level decoder layers focus on learning how to predict accurate location information from blurry edges.

To prove the efficiency of our proposed methods, we conduct comprehensive experiments on the
PIXray dataset and OPIXray dataset. The evaluation results demonstrate that our proposed model is superior to the state-of-the-art object detector.

Our main contributions are summarized as follows:
\begin{enumerate}
    \item We propose a powerful end-to-end object detector for overlapping phenomena in X-ray images, AO-DETR. To our knowledge, this is the first DETR-like model in the field of prohibited item detection. Experimental results show that our model achieved the best performance on multiple datasets.
    \item We propose a CSA strategy, which enhances the anti-overlapping feature extraction capability for specific category foregrounds by constraining the object classes assigned to category-specific queries during the training phase.
    \item The proposed LFD scheme improves the accuracy of reference boxes predicted by mid-level and high-level decoder layers through dense gradient transmission, ultimately enhancing the perception ability of blurry edges of models.
\end{enumerate}

The remainder of this paper is as follows. Initially, Section~\ref{Related Work} provides a review of the relevant methods. This is followed by Section~\ref{Proposed Method}, where we describe our proposed method in detail. Section~\ref{Experiments} then presents and discusses the results of our experiments. The paper concludes with Section~\ref{Conclusions}, summarizing our key findings and observations.
\section{Related Work}\label{Related Work}
We briefly summarize some state-of-the-art object detection methods that have achieved remarkable results. Additionally, we introduce existing methods for enhancing anti-overlapping feature extraction capabilities from two pathways: accurate label assignment and multi-stage regression.
\subsection{Generic Object Detection}

Recently, the object detectors based on convolutional neural networks, such as Faster R-CNN~\cite{Faster}, Cascade R-CNN~\cite{Cascade}, FCOS~\cite{FCOS}, ATSS~\cite{ATSS}, and YOLOX~\cite{YOLOX}, are gradually overtaked by detectors with Detection Transformer framework.
Detection Transformer (DETR)~\cite{DETR} first proposes an end-to-end detection transformer framework while requiring no hand-crafted non-maximum suppression (NMS)~\cite{NMS}. Deformable DETR~\cite{Deformable-DETR} presents the Deformable attention to accelerate convergence of DETR, which combines the advantage of the sparse spatial sampling of deformable convolution, and the global modeling capability of Transformers.
DAB-DETR~\cite{DAB-DETR} presents a new query formulation and using dynamic anchor boxes for DETR.
DN-DETR~\cite{DN-DETR} and DINO~\cite{DINO} enhance DETR-like models from the perspectives of denoising queries and anchor boxes, respectively. $\mathcal{C}$o-DETR~\cite{CO-DETR} combines some traditional label assignment strategies and Hungarian matching to mitigate the sparse supervision problem caused by one-to-one set matching strategy.
\subsection{Label Assignment}

Labeling anchors as positive or negative samples is crucial in detector training, which can mitigate the foreground-background class imbalance problem in X-ray images. Traditional anchor-based detectors like YOLOv3~\cite{YOLOv3}, SSD~\cite{SSD}, Faster R-CNN~\cite{Faster}, and RetinaNet~\cite{RetinaNet} use IoU for label assignment. Anchor-free detectors such as FCOS~\cite{FCOS} and Foveabox~\cite{Foveabox} employ center sampling strategies. However, these methods are sub-optimal due to their fixed rules.
Some advanced Label Assignment strategies \eg, OTA~\cite{OTA}, ATSS~\cite{ATSS}, PAA~\cite{PAA}, and SimOTA~\cite{YOLOX} offer dynamic positive sample selection. OTA~\cite{OTA} views label assignment globally, treating it as an optimal transportation problem. ATSS~\cite{ATSS} uses top-k anchors for threshold determination, while PAA~\cite{PAA} applies a probabilistic method. SimOTA~\cite{YOLOX} focuses on overall cost for better positive sample selection.

Recently, DETR~\cite{DETR} pioneered the use of global matching cost and the Hungarian algorithm to achieve a unique prediction result for each object in images. This represents the first successful application of a one-to-one label assignment scheme.
However, the bipartite graph matching is unstable in the early stage of training phase. DN-DETR~\cite{DN-DETR} introduces noised bounding boxes and labels that bypass the need for Hungarian matching, thereby mitigating the issue of unstable assignments. DINO~\cite{DINO} further put forward contrastive denoising training to accelerate convergence. Group-DETR uses a group-wise one-to-many label assignment, akin to the hybrid matching of $\mathcal{H}$-DETR~\cite{H-DETR}, for multiple positive object queries. Conversely, $\mathcal{C}$o-DETR introduces a collaborative optimization approach for one-to-one set matching, differing from these follow-ups.

Despite all the progress, the role of the learned queries in DETR is still not fully understood or utilized. We are experimenting with enabling queries to specifically target information from certain categories, thereby enhancing their ability to extract features in overlapping scenarios.

\subsection{Multi-stage Regression}
One-stage architectures like the YOLO~\cite{YOLO,YOLOv3,YOLOv7} and SSD~\cite{SSD,DSSD,FSSD} series, designed for real-time performance. These models forgo separate region proposals of the R-CNN framework, processing the image in one pass. Although faster than region-based models, they initially faced challenges with lower detection accuracy.
Two-stage detectors, such as Fast R-CNN~\cite{Fast}, Faster R-CNN~\cite{Faster}, and R-FCN~\cite{R-fcn} utilize a dual-step approach in object detection. Initially, they generate region proposals or candidate bounding boxes likely to contain objects. Subsequently, these proposals are classified and refined for the final detection outcome. 
Cascade R-CNN~\cite{Cascade}, a multi-stage detection system, tackles challenges by using detectors trained with increasing IoU thresholds. This approach allocates different IoU threshold proposals as positive samples at each stage, enabling refinement of regression progressively.
Deformable DETR~\cite{Deformable-DETR} designed a simple iterative mechanism for bounding box refinement to improve detection performance, where each decoder layer refines the boxes based on the output of the prior layer.
To overcome the shortsightedness of refining boxes in each decoder layer, while keeping
the advantages of fast convergence, DINO introduces a novel look forward twice scheme, where the updated parameters are corrected using predictions from the current layer and the next lower layer. 

In this work, we seek to extend this approach, allowing the gradient of the prediction results from the current layer to propagate to the current layer and each lower decoder layer.

\subsection{X-ray Object Detection}
Some CNN-based object detectors~\cite{SIXray,OPIXray,PIXray,Xdet,PIDray,tao2022few,shao2022exploiting,PIXDet,PID-YOLOX} have contributed significantly to security inspection. SIXray~\cite{SIXray} introduces class-balanced hierarchical refinement to address X-ray image complexities and class imbalance. OPIXray~\cite{OPIXray} develops a de-occlusion attention module for X-ray image occlusions, enhancing feature maps with distinct item appearances. PIXray~\cite{PIXray} offers a dataset with diverse prohibited items and proposes a Dense De-overlap Attention Snake for segmentation. Xdet~\cite{Xdet} first statistically analyzes the physical size distribution of different prohibited object categories and found that their physical sizes exhibit clear distinctions. However, GADet~\cite{GADet} points out that the areas of objects obtained by the Otsu~\cite{Otsu} algorithm are not accurate enough in Xdet~\cite{Xdet}, and the area is not as robust as the diagonal length in X-ray images. Thereby, they introduce the Physical Diagonal Length Constraint (PDLC), which can utilize this underlying relationship to align classification and localization tasks in object detection.
Although these methods utilize CNN frameworks, we aim to enhance DETR-like models specifically for X-ray prohibited item detection.

\section{Proposed Method}\label{Proposed Method}
\begin{figure*}
    \centering
    \includegraphics[width=1\linewidth]{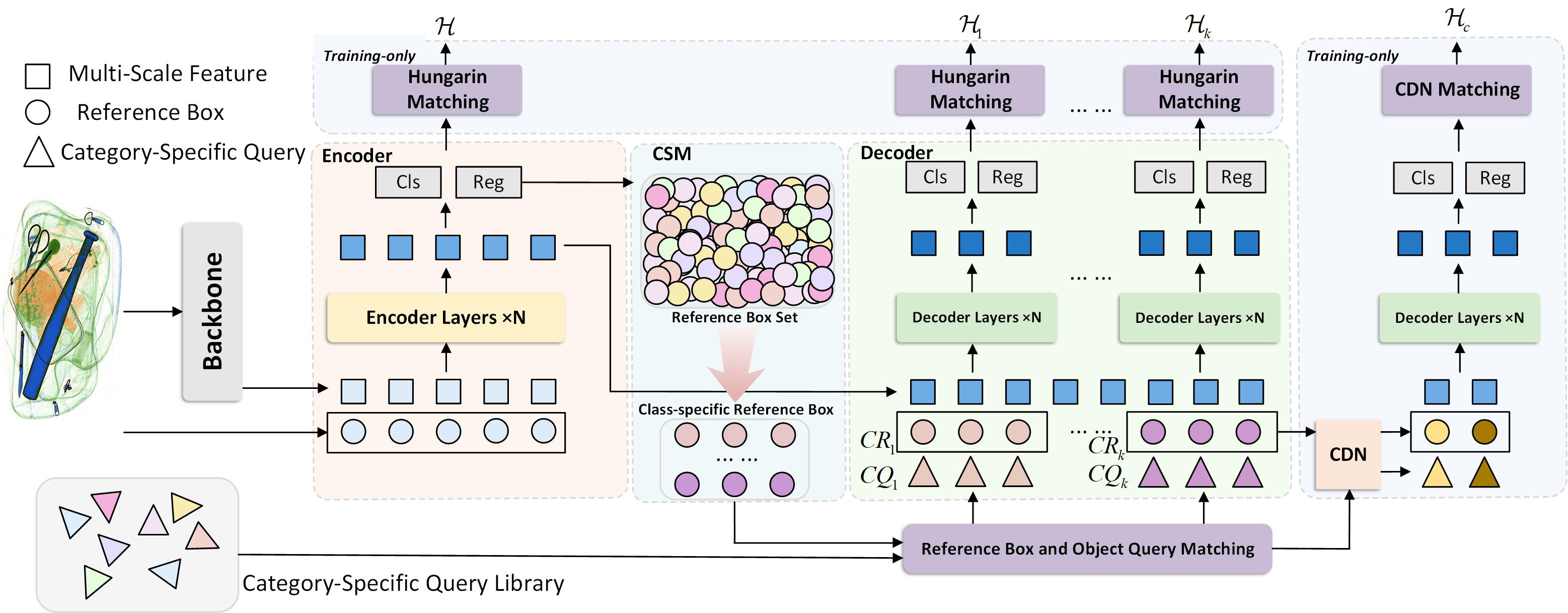}
    \caption{The architecture of AO-DETR. The Backbone, Encoder, Decoder, and CDN modules are the same as DINO~\cite{DINO}. 
    For CSA strategy, we match the category-specific high-quality reference boxes obtained from CSM with their corresponding category-specific object queries before inputting them into the decoder module for prediction. We further employ an additional k-category-specific Hungarian matching mechanism to conduct one-to-one matching on the predicted results. This process serves to enhance the semantic clarity of the object query categories.
    }
    \label{architecture}
\end{figure*}

\subsection{Revisit the Match Part Label Assignment of DINO}

In the DINO framework, the encoder outputs a large number of prediction results $P$, encompassing classification results $C$, and localization results or reference boxes $R$.
Similar to the decoding process of most Deformable DETR series models, they randomly initialize $N_{pred}$ object queries $Q^0$, to guide the corresponding number of reference boxes $R^0$ selected from $R$ to predict objects, where sets $Q^0$ and $R^0$ will be used by the decoder layer 0 for subsequent predictions.
Then, the selection mechanism in DINO picks the top $N_{pred}$ prediction results, denoted as $P^0$, which have the highest classification confidence scores among the set $P$. 
Additionally, they are then assigned one-to-one to each $q^0$, where $q^0\in Q^0$, in descending order of classification confidence. This establishes the pairing relationship between the reference box and the object query for decoder.
For ease of description, we utilize the index $i$ to represent the pairing relationship among reference box, classification confidence, object query, and ground truth. Elements with the same subscript $i$ possess a pairing relationship.
To obtain the $i$-th pair of object query and reference box in layer $l$, where $q_i^l\in Q^l, r_i^l\in R^l$, the decoder procedure of $l$-th layer decoder layer can be expressed in the following form:
\begin{equation}
\label{equation decoder layer i}
q_{i}^l, r_{i}^l, c_i^l= \mathcal{L}^l(q_{i}^{l-1}, r_{i}^{l-1}, X;\theta^l),
\end{equation}
where $l\in \{x \mid x\in \mathbb{Z}, 0< x \leq L\}$, representing the layer index, and $L=6$ denotes the total number of layers. 
$\theta^l$ is the parameters in $l$-th layer, and $X$ denotes the multi-scale features extracted by the encoder.

To establish the one-to-one correspondence between the ground truth labels $G$ and the prediction results $P^l$ from the $l$-th layer of the decoder, it is necessary to compute a matching cost matrix $M\in \mathbb{R}^{N_{pred},N_{gt}}$. The $N_{gt}$ and $N_{pred}$ represent the numbers of ground truths and predictions, respectively. The specific method for computing this cost matrix can be referenced from DETR.
Then, they apply the Hungarian algorithm for bipartite graph matching, and the process can be represented as follows:
\begin{equation}
\label{equation Hungarian all}
\{p_i^l,q_i^l,g_i\mid q\in \mathbb{Z}, 0\leq q<N_{gt}\}=\mathcal{H}^l(P^l,Q^l,G).
\end{equation}
With this step, the algorithm establishes the pairing among the reference boxes, classification confidences, object queries, and ground truths, which can be represented by the subscript '$i$'.

\subsection{Category-Specific One-to-One Assignment}
Although the aforementioned method has achieved significant success, the issue of the unclear categorical significance of queries remains unresolved.
As shown in Fig.~\ref{figure T-SNE}, upon processing X-ray images with significant overlap, the queries in the final layer of the decoder, after being reduced in dimensionality, exhibit a high degree of coupling and are scattered all over the plane. This indicates that the queries have extracted an extensive array of diverse background information and that a single query is unable to extract the features of a category-specific foreground object from the overlapping foreground and background.
We propose a Category-Specific One-to-One Assignment strategy (CSA) to alleviate the issue of unclear categorical significance in queries, while simultaneously enhancing the anti-overlapping capabilities of feature extraction of queries. The four main components of the strategy are as follows.
\subsubsection{Category-Specific Object Queries}
Specifically, we initialize a category-specific object query prototype library at random, denoted as Set $CQ$, each category-specific query prototype $q_k^0 \in CQ$. For the $k$-th category, we define the category-specific query group $Q_k^0$, which contains $N_k$ identical $q^0_{k}$.
Due to the presence of positional encoding and the allocation of different ground truths, each query $q^0_{k,i}$ within the $k$-th query group gradually diverges during the training phase. Since each of them trained from the same prototype, this naturally leads to an effect where queries within the same category become similar, while those from different categories become increasingly dissimilar.
\subsubsection{Category-Specific Select Mechanism}
In decoders with deformable attention, queries guide networks in feature extraction from areas near reference box centers to refine these boxes. Aligning the categories of queries and reference boxes is crucial. 
Queries adept in detecting category A, denoted as $Q_{k=A}$, when paired with reference boxes of category B, denoted as $R_{k=B}$ might lead to network confusion. Specifically, the region within $R_{k=B}$ containing only class B foregrounds and overlapping backgrounds cannot provide the feature of class A for $Q_{k=A}$. Similarly, $Q_{k=A}$ cannot effectively direct network to extract features of class B in region with $R_{k=B}$ for object detection. 
Hence, keeping the category of reference boxes and queries consistent aids the network in extracting relevant foreground features from overlapping backgrounds, enhances the anti-overlapping capability of feature extraction of the model.

In mainstream DETR detectors such as Deformable DETR and DINO, the select strategy directly chooses the top $N$ reference boxes with the highest confidence scores from all encoder predictions and then arranges them in sequence with the object queries in the decoder. This approach, in fact, overlooks the consistency between the categories of the queries and the categories of the reference boxes. To address this issue, we propose a Category-Specific Select Mechanism (CSM) that provides reference boxes corresponding to each category-specific query group, thereby resolving the aforementioned problem of category inconsistency.
The specific process is as illustrated in Algorithm~\ref{algorithm1}. Given all the results $P$ predicted by the last layer of the encoder, the number of categories $K$, and the number of reference boxes $N$ required for all categories, this algorithm can filter out the top $N_k$ predictions with the highest confidence for each category, which are category-specific reference predictions of class $k$, denoted as $P_k^0$. This process can be denoted as follows:
\begin{equation}
\label{equation CAM}
P^0=\{P_k^0 \mid k\in \mathbb{Z}, 0\leq k< K \}=CSM(P,N,K),
\end{equation}
where $P^0$ represents all category-specific prediction results, including category-specific reference boxes $R^0$ and category-specific classification score $C^0$.
\subsubsection{Reference Box and Object Query Matching}
Before decoding, to align the categories of reference boxes and object queries, we perform one-to-one matching between the category-specific reference boxes selected by CSM and category-specific query group with the same category $k$. 
Then, we obtain the $i$-th pair object query $q_{k,i}^0\in Q_k^0$ and reference box $r_{k,i}^0\in R_k^0$, and $i$ represents the matching pair index. This matching relation ensures that a category-specific query for a specific class consistently guides the prediction of the reference box containing the corresponding class throughout the training process.
Furthermore, we obtain the $i$-th pair of object query and reference box in category $k$ for layer $l$, where $q_{k,i}^l\in Q_k^l, r_{k,i}^l\in R_k^l$, and the procedure of $l$-th each decoder layer can be expressed in the following form:
\begin{equation}
\label{equation decoder layer i after CSM}
q_{k,i}^l, r_{k,i}^l, c_{k,i}^l= \mathcal{L}^l(q_{k,i}^{l-1}, r_{k,i}^{l-1}, X;\theta^l),
\end{equation}
if $l=1$, $q_{k,i}^{l-1} \in Q_k^0$ and $r_{k,i}^{l-1} \in R_k^0$, which can be obtained by category-specific query library and category-specific select mechanism.
\subsubsection{Category-Specific Hungarian Matching}
To effectively establish a one-to-one correspondence in category $k$ between the ground truth $G$ and the prediction results of the $l$-th decoder layer $P_k^l$, we compute the category-specific matching cost matrix $M_k\in \mathbb{R}^{N_{k,pred},N_{k,gt}}$. Here, $N_{k,pred}$ and $N_{k,gt}$ represent the number of ground truths and predictions for category $k$, respectively.
Then, we utilize the Hungarian algorithm for bipartite graph matching for each category, and the process can be represented as follows:
\begin{equation}
\label{Hungarian all}
\{p_{k,i}^l,q_{k,i}^l,g_{i}\mid q\in \mathbb{Z}, 0\leq q<N_{gt}\}=\mathcal{H}_k^l(P_k^l,Q_k^l,G).
\end{equation}

At this point, our CSA strategy establishes the pairing among reference boxes, classification scores, object queries, and ground truths in each category $k$.
As shown in Fig.~\ref{figure T-SNE}(c), in T-SNE dimensionality reduction visualization of all category-specific object queries in the last decoder layer of AO-DETR, which utilizes CSA strategy, category-specific queries responsible for gun, scissor, and knife in the last decoder layer converge to distinct corners. They exhibit both intra-class clustering and inter-class repulsion characteristics. This indicates that our CSA strategy has trained queries to extract specific foreground features for particular categories, demonstrating strong anti-overlapping features.



\begin{algorithm}[!ht]
\caption{Category-Specific Select Mechanism.}\label{CSM}
\begin{algorithmic}
\REQUIRE ~~\\ 
$N$ is the number of all queries;\\
$P$ is the prediction results of the last encoder layer, which including the classification scores $C$ and reference boxes $R$;\\
$K$ is the number of categories;

\ENSURE ~~\\ 
build empty set for classification scores: $C^0 \gets \emptyset$;\\
build empty set for reference boxes: $R^0 \gets \emptyset$;\\ 
$N_k=N/K$;\\
\FOR{ $\forall$ category-specific predication results $ P_k\in P$ }
\STATE $C_k^0 \gets$ select $N_k$ reference box $p_{k,p}$ from $P_k$ with highest $c_{k,p}$;
\STATE $I_k \gets$ obtain the index of $c_k^0$ in Set $C_k$;
\STATE $R_k^0 \gets$ get the corresponding reference boxes by $I_k$;
\STATE $C^0=C^0\cup C_k^0 $;
\STATE $R^0=R^0\cup R_k^0 $;
\ENDFOR
\STATE $P^0= R^0 \cup C^0$;
\RETURN $P^0$; 
\end{algorithmic}
\label{algorithm1}
\end{algorithm}
\subsection{Look Forward Densely}
As shown in Fig.~\ref{X-ray images}, due to the presence of overlapping phenomenon, the edges of the firework in subfigure (a), as well as the handle of the screwdriver in subfigure (b), exhibit significant edge blurring phenomenon. Therefore, the overlapping phenomenon in X-ray images leads to issues of inaccurate localization, and precise regression and localization of edges are of critical importance. 
\begin{figure*}
    \centering
    \includegraphics[width=1\linewidth]{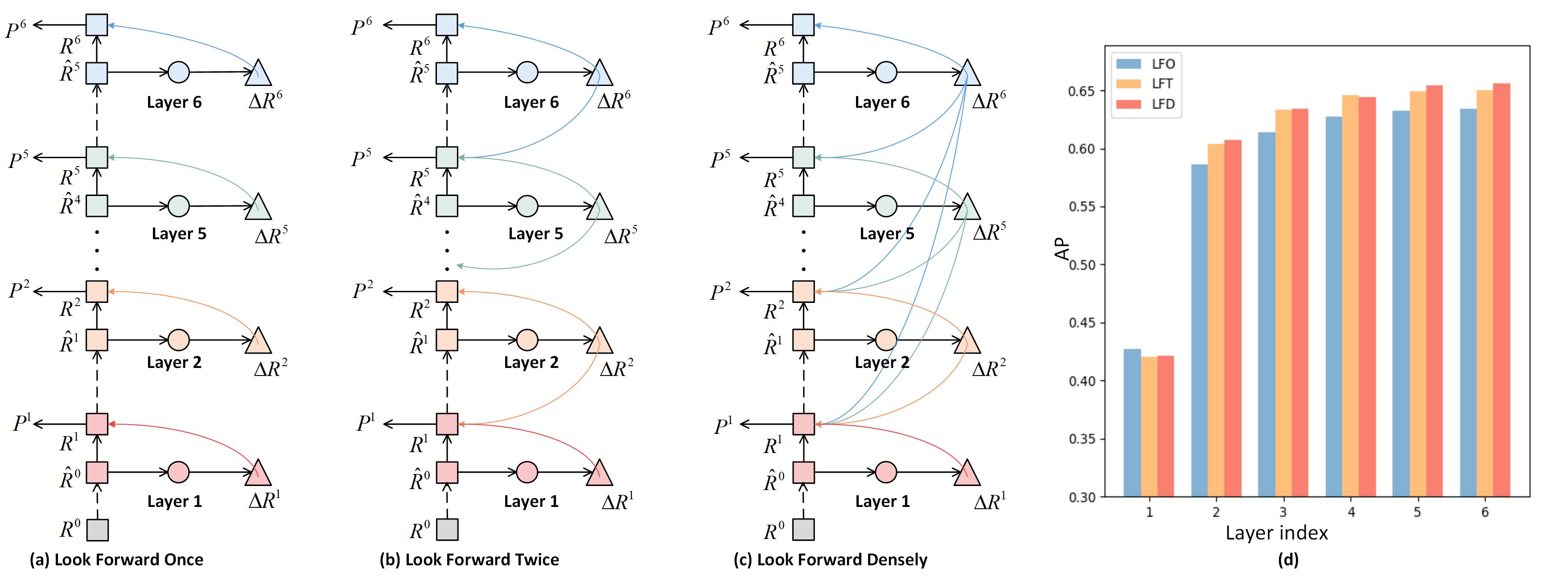}
    \caption{(a)(b)(c) Comparing the structures of Look Forward Once, Look Forward Twice, and Look Forward Densely. (d) APs of look forward
once, look forward twice, and look forward densely in each decoder layer. 'LFO', 'LFT', and 'LFD' are the corresponding abbreviations.
    }
    \label{figure LFD}
\end{figure*}
Deformable DETR~\cite{Deformable-DETR} was the pioneer in introducing an iterative bounding box refinement mechanism in the decoder, assisting the decoder layers in iteratively locating the edges of objects. The changes in the predicted boxes during the process have been shown in subfigure (a) of Fig.~\ref{figure LFD}, and can be represented by the following equations. Given normalized localization boxes $R^{l-1}$ predicted by the $(l-1)$-th decoder, they first stop the gradient as follows.
\begin{equation}
\label{equation LFO 1}
\hat{R}^{l-1}=\mathcal{B}(R^{l-1}),
\end{equation}
where $\mathcal{B}(\cdot)$ represents blocking gradient propagation, a mainstream approach when paired with layer-specific auxiliary loss~\cite{hofinger2020improving}. Then, the obtained $\hat{R}^{l-1}$ are used as the input reference box for the $l$-th decoder layer as follows.
\begin{equation}
\label{equation LFO 2}
\Delta R^{l}=\mathcal{L}_{reg}^l(Q^{l-1},\hat{R}^{l-1},X;\theta^l),
\end{equation}
where $\mathcal{L}_{reg}^l(\cdot)$ represents the regression prediction process at the $l$-th layer. $\Delta R^{l}$ denotes the predicted offset, including $x,y,w,h$~\cite{Deformable-DETR}. Finally, $\hat{R}^{l-1}$ and $\Delta R^{l}$ are utilized for obtaining the normalized localization results of the $l$-th decoder layer, denoted as $R^l$, as follows. 
\begin{equation}
\label{equation LFO 3}
R^l=\sigma (\sigma^{-1}(\hat{R}^{l-1})+\Delta R^{l}),
\end{equation}
where $\sigma(\cdot)$ and $\sigma^{-1}(\cdot)$ denote the sigmoid and inverse sigmoid functions, respectively. It is important to note that this box update approach is designed to ensure that the updated boxes have normalized $x, y, w, h$ values, which range between 0 and 1.
In the aforementioned method, to obtain $l$-th layer reference boxes $R^l$, the deviation predicted by the $l$-th layer $\Delta R^l$ serves only to correct the reference box of $(l-1)$-th layer $\hat{R}^{l-1}$, and the parameters for $l$-th layer $\theta^l$ are updated based exclusively on the corresponding layer-specific auxiliary loss. Consequently, this approach has been termed the Look Forward Once (LFO) strategy by some researchers~\cite{DINO}.
DINO introduces a better Look Forward Twice (LFT) scheme, which utilizes $\Delta R^l$ to guide the prediction processes of both $R^l$ and $R^{l-1}$. LFD scheme tends to degrade the prediction results of the previous layer while improving those of the current layer. This refinement process of reference boxes of layer $l$ can be described as follows:
\begin{equation}
\label{equation LFT}
R^l=\sigma (\sigma^{-1}(\hat{R}^{l-1})+\Delta R^{l}+\Delta R^{l+l}).
\end{equation}

Analyzing the influence of the predicted offset of a single layer that uses the LFT strategy, the offset of the current layer will interfere with the localization results of one lower layer, while directly improving the accuracy of the localization results of the current layer and indirectly improving the accuracy of the localization results of the all higher layers.
Therefore, the current layer with LFT strategy is negatively influenced by one higher layer while positively influenced by one lower layer.
In a holistic view, the lowest layer is only negatively influenced, whereas the highest layer receives only positive guidance. The middle layers are subjected to both positive and negative influences. 
Since the offsets from the higher layers are generally smaller, for one layer, the negative guidance from the higher layer tends to be weaker than the positive guidance from the current layer. As a result, the lower layers suffer, the middle layers benefit marginally, and the upper layers gain significantly. During network inference, the network outputs only the predictions of the highest layer, making LFT strategy advantageous for precise regression at the edges of objects.
However, LFT is relatively conservative. We attempt to further extend this strategy into a more dense form of guidance, termed Look Forward Densely (LFD). The LFD allows the offset predicted by the current layer to participate in the prediction of the reference boxes of all lower layers. The prediction process of reference boxes in layer $l$ can be represented in the following form:
\begin{equation}
\label{equation LFD SE}
R^l_{S,E}=\sigma (\sigma^{-1}(\hat{R}^{l-1})+\sum_{n=l}^L\Delta R^{l}),
\end{equation}
where subscripts '$S$' and '$E$' represent the sum with equaling weighting factor.
The current layer with LFD strategy is negatively influenced by all higher layers while positively influenced by all lower layers.
Compared with LFT, the lowest layer is negatively influenced by other all layers, and the highest layer receives positive guidance from other all layers. The middle layers obtain benefits more times than LFT. 
As a result, compared with LFT, the lower layers suffer more, the middle layers benefit more, and the upper layers gain more. This results in localization by the final layer being more precise, effectively countering the inaccuracies caused by edge blurriness. 
In comparison to the LFT strategy, as shown in Fig.~\ref{figure LFD}(d), our LFD further enhances the predictive outcomes of both the middle and the higher layers. 
Furthermore, we further explore the effects of geometrically scaling the offsets from different layers, either by amplifying or diminishing them in a proportional manner, as demonstrated in the following:

\begin{equation}
\label{equation LFD SA}
R_{S,A}^l=\sigma (\sigma^{-1}(\hat{R}^{l-1})+\sum_{n=l}^L\Delta R^{l}/2^{L-l}),
\end{equation}
\begin{equation}
\label{equation LFD SD}
R^l_{S,D}=\sigma (\sigma^{-1}(\hat{R}^{l-1})+\sum_{n=l}^L\Delta R^{l}/2^l),
\end{equation}
where subscripts '$A$' and '$D$' represent the amplifying and diminishing weighting, respectively.
Finally, we also attempted to apply the averaging operation separately to each of the aforementioned three methods, denoted as $R^l_{V,E}$, $R^l_{V,A}$, and $R^l_{V,D}$, where subscript '$V$' means the average operation.
The most effective form among them is $R^l_{V,D}$. For detailed experimental results, please refer to Section~\ref{Experiments}, where a comprehensive analysis and presentation of the outcomes are provided.
\subsection{Foreground Instability Score}
The Hungarian matching algorithm utilizes the globally optimal solution of the cost matrix for label assignment. However, during the training of the network, distinct ground truth objects may be assigned to a specific query at different epochs. This variability leads to instability in label assignments for queries, manifesting in two aspects: instability in foreground categories assignment and instability in foreground objects assignment.
In order to quantitatively evaluate the instability of foreground label assignment, we designed a metric named foreground instability score (FIS) as follows. For one training image, decoders of model predict objects $P^j=\{P_0^j,P_1^j,...,P_{N_{pred}-1}^j\}$, where $N_{pred}$ represents the number of predicted objects, and the ground truth objects are denoted as $G=\{G_0^j,G_1^j,...,G_{N_{gt}-1}^j\}$, where $N_{gt}$ represents the number of ground truth objects. After label assignment, we compute an index vector $V^j=\{V_0^j,V_1^j,...,V_{N_{pred}-1}^j\}$ to store the ground truth object assignment results for epoch $j$ as follows:
\begin{equation}
\label{equation v}
V_n^j =
\begin{cases}
m,  & \text{if } P_n^j \text{ matches } G_m. \\
-1, & \text{if } P_n^j \text{ matches nothing}.
\end{cases}
\end{equation}
Similar, we compute an index vector $T^j=\{T_0^j,T_1^j,...,T_{N_{pred}-1}^j\}$ to store the ground truth object assignment results for epoch $j$ as follows:
\begin{equation}
\label{equation t}
T_n^j =
\begin{cases}
c,  & \text{if } P_n^j \text{ matches } c\text{-th category } G_m. \\
-1, & \text{if } P_n^j \text{ matches nothing}.
\end{cases}
\end{equation}
We define the foreground category instability of epoch $j$ for one training image as the difference between its $T_0^j$ and $T_1^{j-1}$ as follows:
\begin{equation}
\label{equation FCS}
FCS^j=\sum_{n=0}^{N_{pred}}\mathbbm{1}(T_n^j\neq T_n^{j-1})\cdot \mathbbm{1}(T_n^j\neq -1 \wedge T_n^{j-1}\neq -1),
\end{equation}
where $\mathbbm{1}(x)$ is 1 if x is true and 0 otherwise, and the symbol '$\wedge$' represents logical AND. $\mathbbm{1}(T_n^j\neq -1 \wedge T_n^{j-1}\neq -1)=1$ means that $T_n^{j}$ and $T_n^{j-1}$ are both responsible for foreground objects.
Similar, we define the foreground objects instability of epoch $j$ for one training image as the difference between its $V_0^j$ and $V_1^{j-1}$ as follows:
\begin{equation}
\label{equation FOS}
FOS^j=\sum_{n=0}^{N_{pred}}\mathbbm{1}(V_n^j\neq V_n^{j-1})\cdot \mathbbm{1}(V_n^j\neq -1 \wedge V_n^{j-1}\neq -1).
\end{equation}
Finally, we take the average of both and normalize it by the number of predicted objects $N_{pred}$.
\begin{equation}
\label{equation FIS}
FIS^j=\frac{FCS^j+FOS^j}{2\cdot N_{pred}}.
\end{equation}
The instability of epoch $j$ for the entire dataset is averaged over the instability numbers for all images. 
We omit the image index for notation simplicity in Equations~\ref{equation v}-\ref{equation FIS}.

FIS comprehensively considers the instability of both foreground category assignments and object assignments, the lower the FIS value, the more stable the label assignment between object queries and foreground objects. 
In Section~\ref{Experiments}, we will analyze the impact of CSA on the model training process using our FIS metric and the IS metric~\cite{DN-DETR}.

\section{Experiments}\label{Experiments}
In this section, we first conduct comprehensive ablation experiments on ResNet-50~\cite{ResNet} and Swin-L~\cite{Swin-Transformer} backbone networks to analyze the compatibility of CSA and LFD.
Then, we analyze the impact of CSA on the model training process using our FIS metric and the IS metric~\cite{DN-DETR}.
Subsequently, we explore six schemes of dense guidance in LFD. Finally, we design extensive experiments on the PIXray and OPIXray datasets to assess the performance of our model. We compare our model with state-of-the-art models from both general detectors and prohibited items detectors in a unified environment.
Furthermore, we commence with a visual analysis of the sampling points of the decoder layer in AO-DETR equipped with CSA.
Finally, we visualize the results of baseline on images with severe overlapping phenomena and draw conclusions.
\subsection{Experimental Setup}
 \subsubsection{Datasets and Evaluation Metrics}
For the PIXray dataset~\cite{PIXray}, we utilize the COCO~\cite{COCO} evaluation metrics. The primary challenge metric is the mean average precision (AP), computed across 10 Intersection over Union (IoU) thresholds ranging from 0.5 to 0.95 with a step size of 0.05. AP$_{50}$ represents the mean average precision calculated at a single IoU threshold of 0.5, while AP$_{75}$ represents the mean average precision at a single IoU threshold of 0.75. Additionally, AP$_{S}$, AP$_{M}$, and AP$_{L}$ represent the AP for small objects ($\text{area} < 32^2$), medium objects ($32^2 < \text{area} < 96^2$), and large objects ($96^2 < \text{area}$), respectively.

For OPIXray, we adopt the VOC~\cite{VOC} evaluation metric. AP is calculated from the area under the Precision-Recall curve of one category at the IoU threshold of 0.5. The mean average precision (mAP) is then computed as the average of AP of all categories. mAP serves as a comprehensive evaluation criterion, effectively representing the accuracy and recall of the detector. It provides a holistic assessment that captures the performance strengths and weaknesses of the detector.
\subsubsection{Implementation Details}
For the sake of fair comparison, we ensure that all models are trained under identical conditions. Each model utilizes the ImageNet~\cite{ImageNet} pre-trained model, including ResNet-50~\cite{ResNet}, ResNeXt-101~\cite{ResNeXt}, and Swin-L~\cite{Swin-Transformer}. The convolutional models employ the SGD optimizer with a learning rate of 0.01, momentum of 0.9, and weight decay of 0.1, while transformer models such as DINO utilize the AdamW optimizer with a learning rate of 0.0001, and weight decay of 0.0001. All models undergo 12 training epochs and are implemented based on the MMDetection framework. For warm-up scheme of convolutional models, during the initial 500 iterations, the learning rate gradually increases in a linear fashion with a warming-up ratio of 0.001. Following this warm-up phase, the learning rate undergoes a step-wise decrement, with specific adjustments occurring at the 8-th and 11-th epoch. For DETR-like models, following Deformable DETR~\cite{Deformable-DETR}, the learning rate is decayed at the 11-th epoch by a factor of 0.1.
 Our AO-DETR utilizes $l_1$ and GIoU~\cite{GIoU} losses for box regression and Quality Focal Loss~\cite{GFLv1} for classification.
 All training is conducted on a uniform computer platform equipped with an NVIDIA GeForce RTX 4090 GPU, an Intel Core i9-13900K CPU, 64GB memory, Windows 10 system, and PyTorch 1.13.1.
\subsection{Ablation Study}
\subsubsection{Ablation Study of AO-DETR}
As shown in Table~\ref{experiment ablation study of CSA and LFD }, we conducted ablation experiments on the PIXray dataset using ResNet-50 and Swin-L backbones, respectively. These experiments aimed to analyze the impact of our CSA and LFD on the baseline detection performance and assess the compatibility between the two methods. CSA and LFD respectively improve the AP of DINO with ResNet-50 from $64.3\%$ to $65.0\%$ and $65.2\%$, which demonstrates their effectiveness.
Simultaneously integrating both CSA and LFD, AO-DETR achieves the highest AP of $65.6\%$, AP$_{S}$ of $23.9\%$, AP$_{M}$ of $50.7\%$, and AP$_{L}$ of $74.8\%$. This underscores the exceptional performance of AO-DETR and highlights the complementary nature of CSA and LFD.
\begin{figure}
    \centering
    \includegraphics[width=1\linewidth]{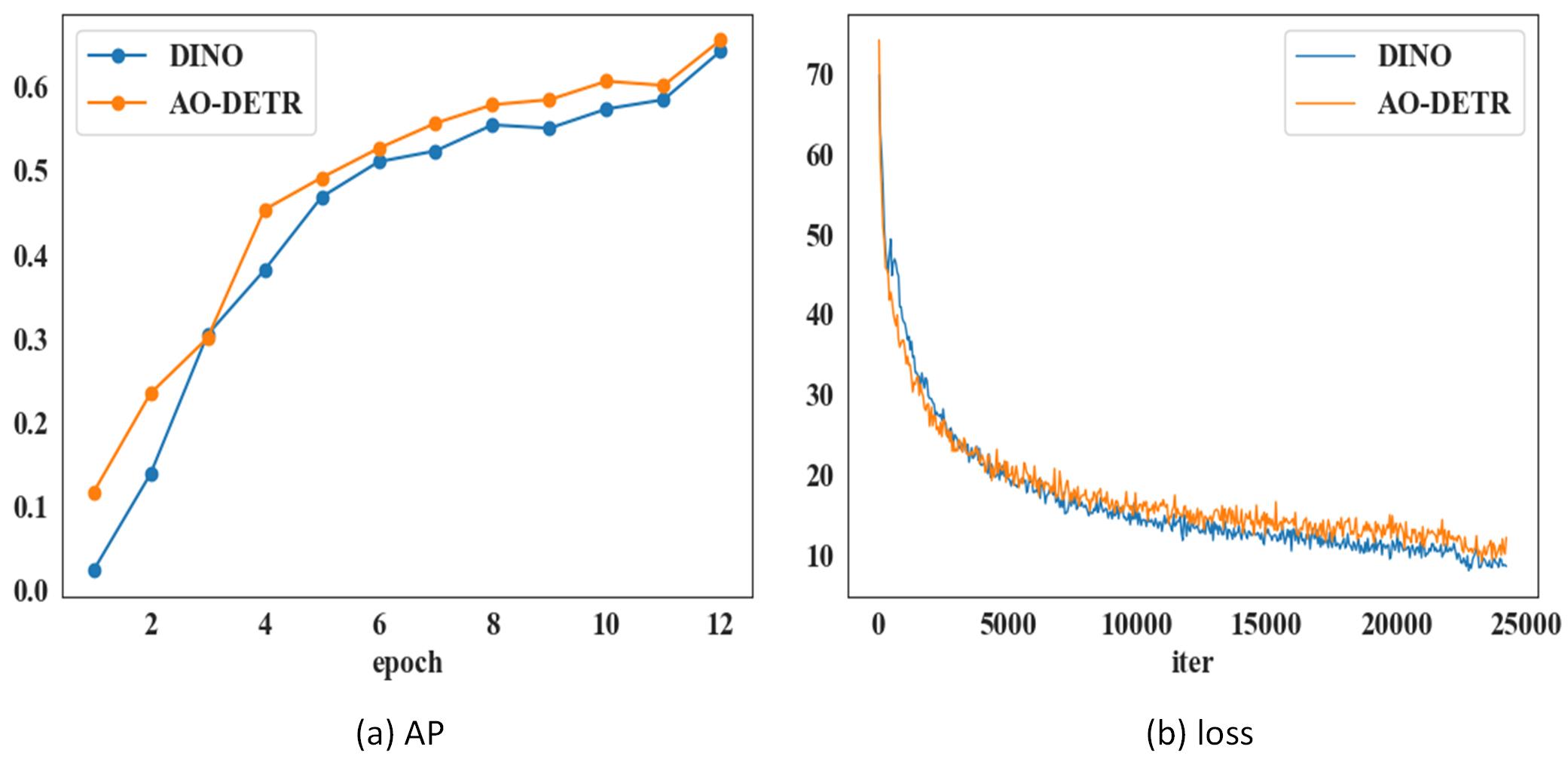}
    \caption{
    (a) The AP curve of DINO~\cite{DINO} and AO-DETR on PIXray~\cite{PIXray} dataset. (b) The loss convergence curve of DINO and AO-DETR on PIXray~\cite{PIXray} dataset.
    }
    \label{figure convergence analysis}
\end{figure}
  \begin{table}[!ht]
      \caption{Ablation results of the proposed CSA and LFD on the PIXray~\cite{PIXray} dataset. “CSA” and “LFD” respectively represent the proposed category-specific one-to-one assignment strategy and look forward densely scheme. }
     \centering
     \resizebox{\linewidth}{!}{
    \large
    \begin{tabular}{c|cc|ccccccc}
         \hline
         Backbone &CSA&LFD&AP & AP$_{50}$ & AP$_{75}$ & AP$_{S}$ & AP$_{M}$ & AP$_{L}$\\
        \hline
    \multirow{4}*{ResNet-50}
           &\XSolidBrush&\XSolidBrush&64.3&86.5&71.0&19.3&48.9&73.9\\
        ~&\Checkmark&\XSolidBrush&65.0&86.1&\textbf{72.7}&22.8&50.1& 74.0\\
        ~&\XSolidBrush&\Checkmark&65.2&\textbf{86.7}&71.5&23.7&49.2&74.5\\
        ~&\Checkmark&\Checkmark&\textbf{65.6}&86.1&72.0&\textbf{23.9}&\textbf{50.7}&\textbf{74.8}\\
            \hline
    \multirow{4}*{Swin-L}
           &\XSolidBrush&\XSolidBrush&72.8 &90.0&80.1&38.3&60.4 &80.4\\
        ~&\Checkmark&\XSolidBrush&73.2&90.3&79.9&38.6&62.4&80.6\\
        ~&\XSolidBrush&\Checkmark&73.4&\textbf{90.6}&79.4&39.8&61.6&81.5\\
        ~&\Checkmark&\Checkmark&\textbf{73.9} &89.9 &\textbf{80.6}&\textbf{40.5}&\textbf{62.4}&\textbf{81.6}\\
        \hline
     \end{tabular}
     }
     \label{experiment ablation study of CSA and LFD }
 \end{table}
Additionally, we also compare the AP and loss convergence curves of AO-DETR and DINO, as shown in Fig.~\ref{figure convergence analysis}. In terms of the AP curve, AO-DETR consistently outperforms the baseline throughout the entire process. Regarding the loss curve, AO-DETR demonstrates a faster convergence in the early stages and stable convergence in the later stages.
\subsubsection{Ablation Study of CSA}
\begin{figure}
    \centering
    \includegraphics[width=1\linewidth]{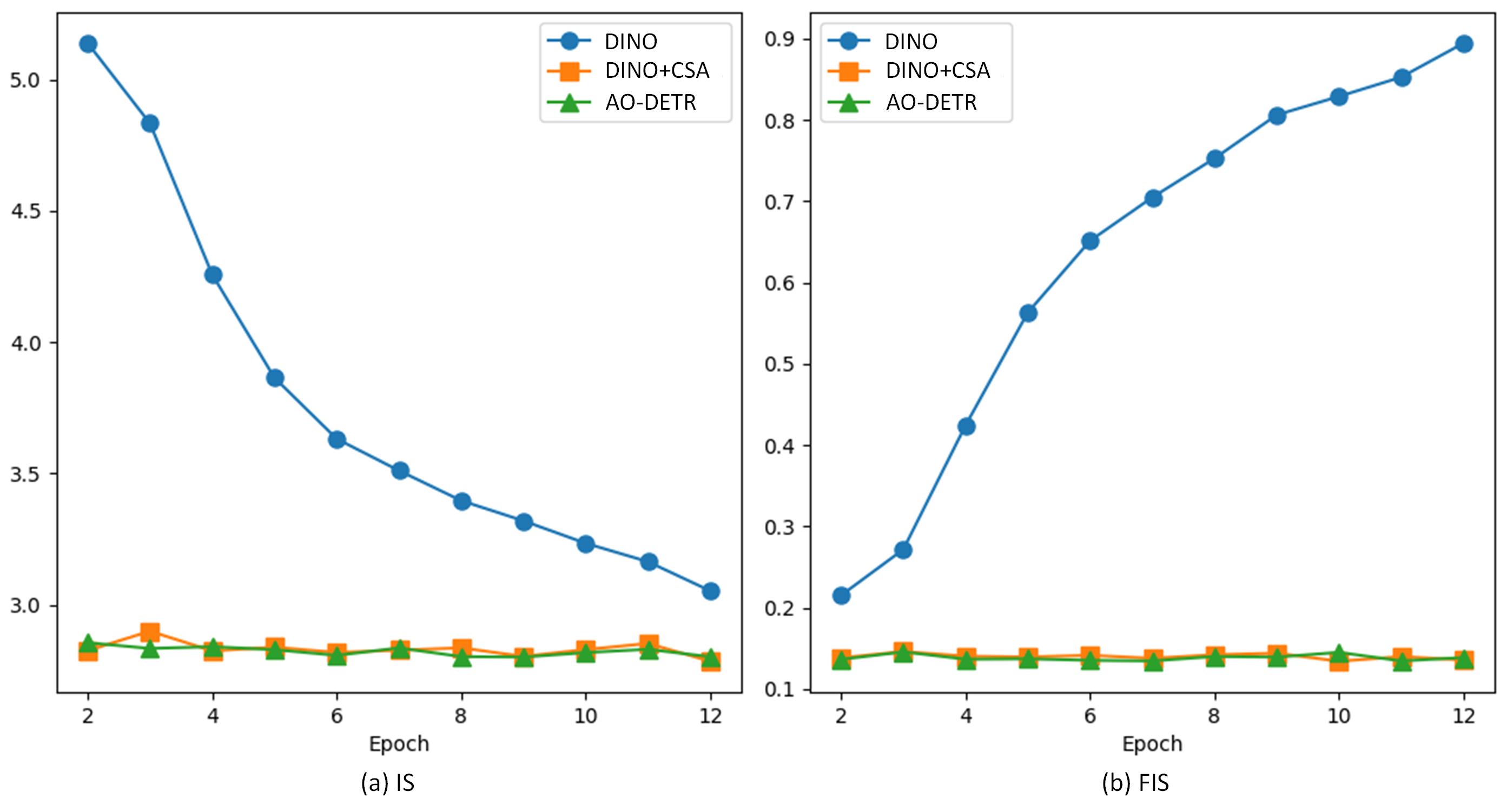}
    \caption{(a) The instability score (IS)~\cite{DN-DETR} of DINO, DINO+CSA, and DINO+CSA+LFD (AO-DETR) on PIXray~\cite{PIXray} dataset. (b) The foreground instability score (FIS) of them. 
    }
    \label{figure FIS}
\end{figure}
To comprehensively analyze the impact of CSA on the model training process, we analyze the instability of models during the training process using the IS metric~\cite{DN-DETR} in conjunction with our FIS metric.
IS~\cite{DN-DETR} is an indicator that evaluates the instability of object assignments without considering whether the predicted results of object queries are allocated to background objects. The lower the IS value, the more stable the label assignment between object queries and all objects including foreground and background.
As illustrated in Fig.~\ref{figure FIS}, with the training of DINO, the IS value decreases. This is attributed to the network's improved ability to discern whether a query is suitable for foreground object responsibility, reducing the probability of erroneously assigning low-capacity queries to predict foreground objects. However, due to the inherent randomness in the original Hungarian matching method, queries are uncertain about their responsibility for a specific category or object in the foreground. This confusion deepens as the network undergoes training, ultimately leading to an increase in the FIS metric.

In contrast, CSA alleviates this confusion significantly by constraining the object classes assigned to category-specific queries during the training phase. Therefore, with the incorporation of CSA, this confusion faced by object queries is greatly mitigated, and both IS and FIS metrics consistently maintain at lower levels. As the object queries become responsible for fixed categories, they can specialize in the features of objects belonging to those categories, thereby enhancing the anti-overlapping feature extraction capability of network.
\subsubsection{Ablation Study of LFD}
   \begin{table}[!ht]
      \caption{Ablations for different types of LFD strategy on the PIXray~\cite{PIXray} dataset.}
     \centering
\begin{tabular}{l|ccc|ccc}
        \hline
Type&AP & AP$_{50}$ & AP$_{75}$ & AP$_{S}$ & AP$_{M}$ & AP$_{L}$\\
        \hline
$R^l$&64.3&86.5&71.0&19.3&48.9&73.9\\
\hline
$R^l_{S,E}$&64.8&86.2&71.7&21.6&50.1&74.0\\
$R^l_{S,A}$&63.9&85.5&70.8&20.2&48.2&73.6\\
$R^l_{S,D}$&64.9&86.4&71.8&22.9&49.9&73.8\\
\hline
$R^l_{V,E}$&65.0&86.3&72.5&22.3&\textbf{50.6}&74.3\\
$R^l_{V,A}$&64.5&85.4&72.1&22.1&49.5&74.2\\
$R^l_{V,D}$&\textbf{65.2}&\textbf{86.7}&\textbf{72.8}&\textbf{23.7}&50.2&\textbf{74.5}\\
            \hline
     \end{tabular}
     \label{experiment ablation study for LFD }
 \end{table}
 To fully explore the potential of LFD scheme, we design and compare six different dense guidance terms, employing equaling (E), amplifying (A), and diminishing (D) strategies in both averaging (V) and summing (S) modes. The specific formulas can be referred to in Equations~\ref{equation LFD SE} to~\ref{equation LFD SD}. 
 As shown in Table~\ref{experiment ablation study for LFD }, we initially assess the impact of employing equaling, amplifying, and diminishing strategies under the summing strategy on the average precision (AP). We observe that $R^l_{S,D}$ achieves a superior AP of $64.9\%$ compared to $R^l_{S,A}$ and $R^l_{S,E}$ with $63.9\%$ and $64.8\%$, respectively. Subsequently, under the averaging strategy, we compare the average precision for employing equaling, amplifying, and diminishing strategies. Here, the AP of $R^l_{V,D}$ at $65.2\%$ outperforms the AP of $R^l_{V,A}$ and $R^l_{V,E}$, which are $64.8\%$ and $64.5\%$, respectively. In summary, "V" is superior to "S", "D" is superior to both "A" and "E", and the overall performance of mode $R^l_{V,D}$ is the best, demonstrating the highest detection accuracy.

 \begin{table*}[ht]
      \caption{Comparison with state-of-the-art general detectors on PIXray~\cite{PIXray}. BS, PARAMs, GFLOPs, and FPS represent batch size, the total number of parameters, the Giga Floating Point operations, and the number of inferences the model can perform per second, respectively.}
     \centering

     \begin{tabular}{l|l|c|c|ccc|ccc|cc|c}
         \hline
          Method & Backbone & BS & FPS &AP & AP$_{50}$ & AP$_{75}$ & AP$_{S}$ & AP$_{M}$ & AP$_{L}$ &PARAMs &GFLOPs& Reference \\
        \hline
    Faster R-CNN~\cite{Faster}& ResNet-50& 16& 87&51.2&79.9&57.6& 3.1&35.4 &60.1&41.19 M& 20.36& TPAMI17\\
    Faster R-CNN~\cite{Faster}& ResNeXt-101-32x4d& 16& 70&53.6& 82.3&60.8& 3.9& 37.7& 62.7&59.83 M& 28.35& TPAMI17 \\

    Mask R-CNN~\cite{Mask} & ResNet-50&16 & 86 &50.0&79.9&56.2&3.8&34.8&59.1&41.42 M& 20.36&ICCV17\\
    Mask R-CNN~\cite{Mask} & ResNeXt-101-32x4d&16 & 73&52.4&81.9&59.4&4.2&36.2&61.3&60.04 M& 28.35& ICCV17\\
    Cascade R-CNN~\cite{Cascade}& ResNet-50& 16& 61&56.5& 81.3& 63.2&8.0& 41.0& 65.9&68.97 M& 22.37& CVPR18\\
    ATSS~\cite{ATSS}& ResNet-101& 16& 66 &52.8&80.8&60.2&7.0&37.4&63.6&51.14 M& 27.82&CVPR20\\
    GFLv1~\cite{GFLv1}& ResNeXt-101-32x4d&16& 66&57.5&82.8&66.0&9.1&42.0&67.4& 50.70 M& 28.51& NeurIPS20\\
    Deformable DETR~\cite{Deformable-DETR}&ResNet-50&2&60&44.6&74.2&48.5&9.6&30.0&53.0&52.14 M&13.47&ICLR21\\
    DINO~\cite{DINO}& ResNet-50&2 &54&64.3&86.5&71.0&19.3&48.9&73.9& 58.38 M&26.89&ICLR23\\
    DINO~\cite{DINO}& Swin-L&2&40&72.8 &\textbf{90.0}&80.1&38.3&60.4 &80.4& 229.0 M&156.0&ICLR23\\
            \hline
    AO-DETR \textbf{(ours)}& ResNet-50&2&54 &65.6&86.1&72.0&23.9&50.7&74.8 & 58.38 M&26.89&---\\
    AO-DETR \textbf{(ours)}& Swin-L&2&40 &\textbf{73.9} &89.9 &\textbf{80.6}&\textbf{40.5}&\textbf{62.4}&\textbf{81.6}& 229.0 M&156.0&---\\
            \hline
     \end{tabular}
     \label{experiment SOTA comparison on PIXray}
 \end{table*}

 \begin{table*}
      \caption{Comparison with state-of-the-art general detectors on OPIXray~\cite{OPIXray}. FO, ST, SC, UT, and MU represent Folding Knife, Straight Knife, Scissor, Utility Knife, and Multi-tool Knife, respectively}
     \centering
     \begin{tabular}{l|l|c|c|c|ccccc|cc|c}
         \hline
          Method & Backbone & BS & FPS &mAP &FO& ST& SC& UT& MU  &PARAMs &GFLOPs& Reference \\
        \hline
    Faster R-CNN~\cite{Faster}& ResNet-50& 16& 87&70.9&75.1&45.8&88.4&67.7&77.3&41.19 M& 20.36& TPAMI17\\
    Faster R-CNN~\cite{Faster}& ResNeXt-101-32x4d& 16& 70&73.4&80.6&45.4&89.1&69.1&83.1&59.83 M& 28.35& TPAMI17 \\

    Mask R-CNN~\cite{Mask} & ResNet-50&16 & 86 &74.7&77.9&51.4&89.5&69.4&85.5&41.42 M& 20.36&ICCV17\\
    Mask R-CNN~\cite{Mask} & ResNeXt-101-32x4d&16 & 73&77.2&83.6&55.9&89.8&71.5&85.2&60.04 M& 28.35& ICCV17\\
    Cascade R-CNN~\cite{Cascade}& ResNet-50& 16& 61&72.8 &75.7& 50.0& 89.4& 70.0 &79.0 &68.97 M& 22.37& CVPR18\\
    ATSS~\cite{ATSS}& ResNet-101& 16& 66 &67.5& 72.8& 38.0 &88.6& 58.0& 80.2&51.14 M& 27.82&CVPR20\\
    GFLv1~\cite{GFLv1}& ResNeXt-101-32x4d&16& 66& 75.6& 80.0 &53.6& 89.3 &71.7& 83.4&50.70 M& 28.51& NeurIPS20\\
    Deformable DETR~\cite{Deformable-DETR}&ResNet-50&2&60&63.4& 70.1& 29.0 &86.0& 55.7& 76.4&52.14 M&13.47&ICLR21\\
    DINO~\cite{DINO}& ResNet-50&2 &54&78.2&83.2&58.8&89.4&72.7&86.7& 58.38 M&26.89&ICLR23\\
    DINO~\cite{DINO}& Swin-L&2&40&80.0& 84.2& 61.1& 89.0 &\textbf{78.9}& 86.6& 229.0 M&156.0&ICLR23\\
            \hline
    AO-DETR \textbf{(ours)}& ResNet-50&2&54 &79.2&83.8&60.5&90.1&74.7&87.1&   58.38 M&26.89&---\\
    AO-DETR \textbf{(ours)}& Swin-L&2&40&\textbf{80.8}&\textbf{84.8}&\textbf{63.0}&\textbf{90.1}&77.7&\textbf{88.4}& 229.0 M&156.0&---\\
            \hline
     \end{tabular}
     \label{experiment SOTA comparison on OPIXray}
 \end{table*}

   \begin{table*}
      \caption{Comparison with state-of-the-art prohibited item detectors on OPIXray~\cite{OPIXray}. IJON stands for the journal Neurocomputing. MAX represents the model trained until it no longer converges}
     \centering
     \begin{tabular}{l|c|c|c|c|ccccc|c}
         \hline
          Method & Backbone&Epoch&Image Size &mAP &FO& ST& SC& UT& MU&Reference \\
        \hline
    DOAM~\cite{OPIXray}&ResNet-50&MAX&--&82.41& 86.71 &68.58 &90.23& 78.84& 87.67&ACM MM20\\
    FA~\cite{FA}&ResNet-50&30&-- &85.8& 89.8& 81.0& 79.6& \textbf{88.6}& 89.9&ICMLA21\\

    XDet~\cite{Xdet}&ResNet-50&MAX&1280&86.69 &90.42 &75.95 &91.46 &84.31 &91.29&KBS22 \\

    ATSS+LAreg~\cite{CLCXray}& ResNet-50&12&1280&87.39&\textbf{92.78}&71.17&96.61&83.45&92.92&TIFS22\\
    ATSS+LAcls~\cite{CLCXray}& ResNet-50&12&1280&88.26&90.04&74.99&97.60&85.70&\textbf{92.96}&TIFS22\\
    MCIA-FPN~\cite{MCIA-FPN}&ResNet-101&MAX&--&82.59&89.08&74.48&89.99&86.13&89.75&TVC23\\
    POD-F-R~\cite{POD}&ResNet-50&24&1333&84.9&88.7&76.0&88.9&82.8&88.1&IJON23\\
    POD-F-X~\cite{POD}&ResNeXt-50&24&1333&86.1&89.4&78.7&90.6&83.3&88.7&IJON23\\
    GADet-S~\cite{GADet}& Modified CSP v5 &60&320&69.6& 72.6& 43.6&86.6& 67.5& 77.5&JSEN24\\
    GADet-L~\cite{GADet}& Modified CSP v5 &60&320&77.7& 81.8& 54.0&89.8& 77.5& 85.2&JSEN24  \\
    GADet-X~\cite{GADet}& Modified CSP v5 &60&320&78.1& 83.1& 56.3&89.8& 75.7& 85.5&JSEN24\\
        \hline
    AO-DETR \textbf{(ours)}& ResNet-50&15&640&87.2& 90.0& 80.1& 90.8& 85.6 &89.5&---\\
    AO-DETR \textbf{(ours)}& Swin-L&15&640&\textbf{89.0}& 89.4& \textbf{80.4} &\textbf{97.8}& 87.4& 90.0&---\\
            \hline
     \end{tabular}
     \label{experiment SOTA prohibited item detector comparison on OPIXray}
 \end{table*}
\subsection{Comparison with State-of-the-Art Methods}
\subsubsection{Comparison with General Detectors}
To validate the superior performance of the proposed model for X-ray image prohibited item detection compared to general object detectors, we compare AO-DETR with ResNet-50 and AO-DETR with Swin-L against state-of-the-art detectors in mainstream prohibited item datasets PIXray~\cite{PIXray} and OPIXray~\cite{OPIXray}, as well as general-domain detectors. These general detectors include models based on convolutional neural networks, such as Faster R-CNN~\cite{Faster}, Mask R-CNN~\cite{Mask}, Cascade R-CNN~\cite{Cascade}, GFLv1~\cite{GFLv1}, ATSS~\cite{ATSS}, and models based on Transformers, such as Deformable DETR and DINO. For fairness, all models are trained for 12 epochs, which are all reimplemented by MMDetection~\cite{MMDetection}, and images are resized to $320\times 320$. 
The number of object queries in DINO and category-specific object queries of our AO-DETR remains consistent, both being twice the number of categories.
As shown in Table~\ref{experiment SOTA comparison on PIXray}, our AO-DETR with Swin-L achieves the highest detection accuracy on the PIXray dataset, with an AP of $73.9\%$, surpassing other general detectors significantly. To balance real-time requirements, we introduce a lightweight version, AO-DETR with ResNet-50, reducing PARAMS from 229 M to 58.38 M, GFLOPs from 156 G to 26.89 G, and improving frame rate from 40 frames per second to 54 frames per second.
The AP remains high at $65.6\%$, surpassing the best general detector GFLv1 at $57.5\%$, with the exception of DINO.
Furthermore, AO-DETR, compared to the baseline model DINO, shows no increase in required GFLOPs and PARAMs during the inference process, and there is no decrease in FPS. This indicates that our approach does not require additional computational resources during inference. While maintaining inference speed, the detection accuracy of the smaller model improves from $64.3\%$ and $72.8\%$ to $65.6\%$ and $73.9\%$, respectively.

To demonstrate the robustness of our method, we perform a comparative analysis on another mainstream prohibited item detection dataset, OPIXray. The results, as shown in Table~\ref{experiment SOTA comparison on OPIXray}, align with the conclusions mentioned above, indicating the consistent performance of AO-DETR across different datasets.
 \subsubsection{Comparison with Prohibited Items Detectors}
 To validate the superiority of AO-DETR over other prohibited item detectors, we conducted comparisons on the OPIXray dataset. In order to challenge the performance limits of AO-DETR, we extend the number of training epochs from 12 to 15, increase the object query quantity from the default twice the number of categories to twenty times the number of categories, and enlarge the image input size from $320\times320$ to $640\times640$. The relevant parameters for other prohibited item detectors were referenced from their respective papers.
 The results, as shown in Table~\ref{experiment SOTA prohibited item detector comparison on OPIXray}, demonstrate that our AO-DETR with Swin-L detector, trained on lower-resolution images for only 15 epochs, achieves a mean average precision (mAP) surpassing other state-of-the-art models trained at higher resolutions, such as POD-F and MCIA-FPN.
\begin{figure*}
    \centering
    \includegraphics[width=1\linewidth]{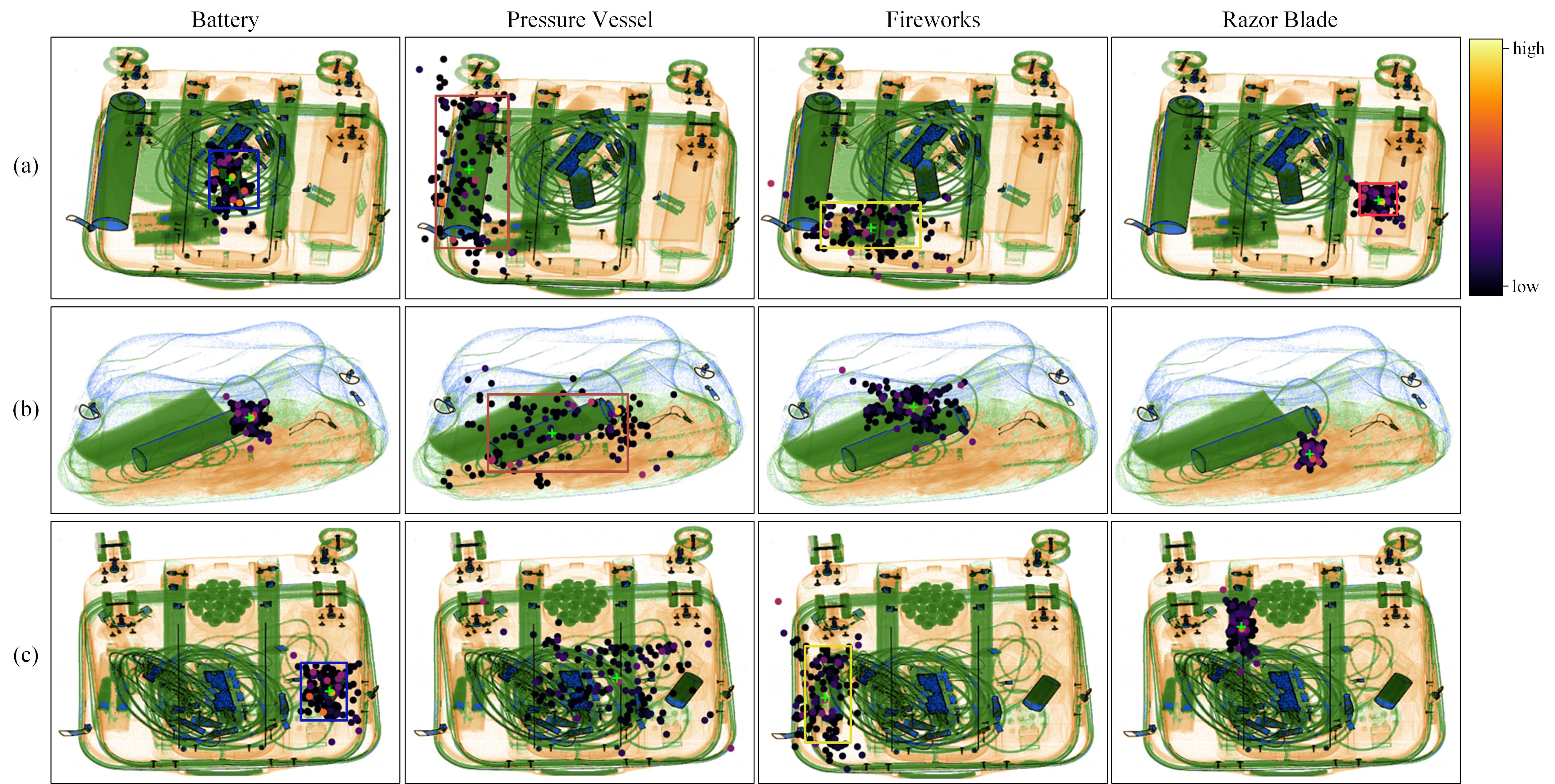}
    \caption{
Visualization of deformable attention sampling points and reference points of corresponding category-specific object queries in the last decoder layer. Each sampling point is depicted as a filled circle, with its color reflecting its corresponding attention weight. The reference point is represented by a green cross marker.
The predicted bounding boxes whose confidence scores are over threshold value have been shown with category-specific color.
    }
    \label{figure sampling points and reference points}
\end{figure*}
\begin{figure*}
    \centering
    \includegraphics[width=1\linewidth]{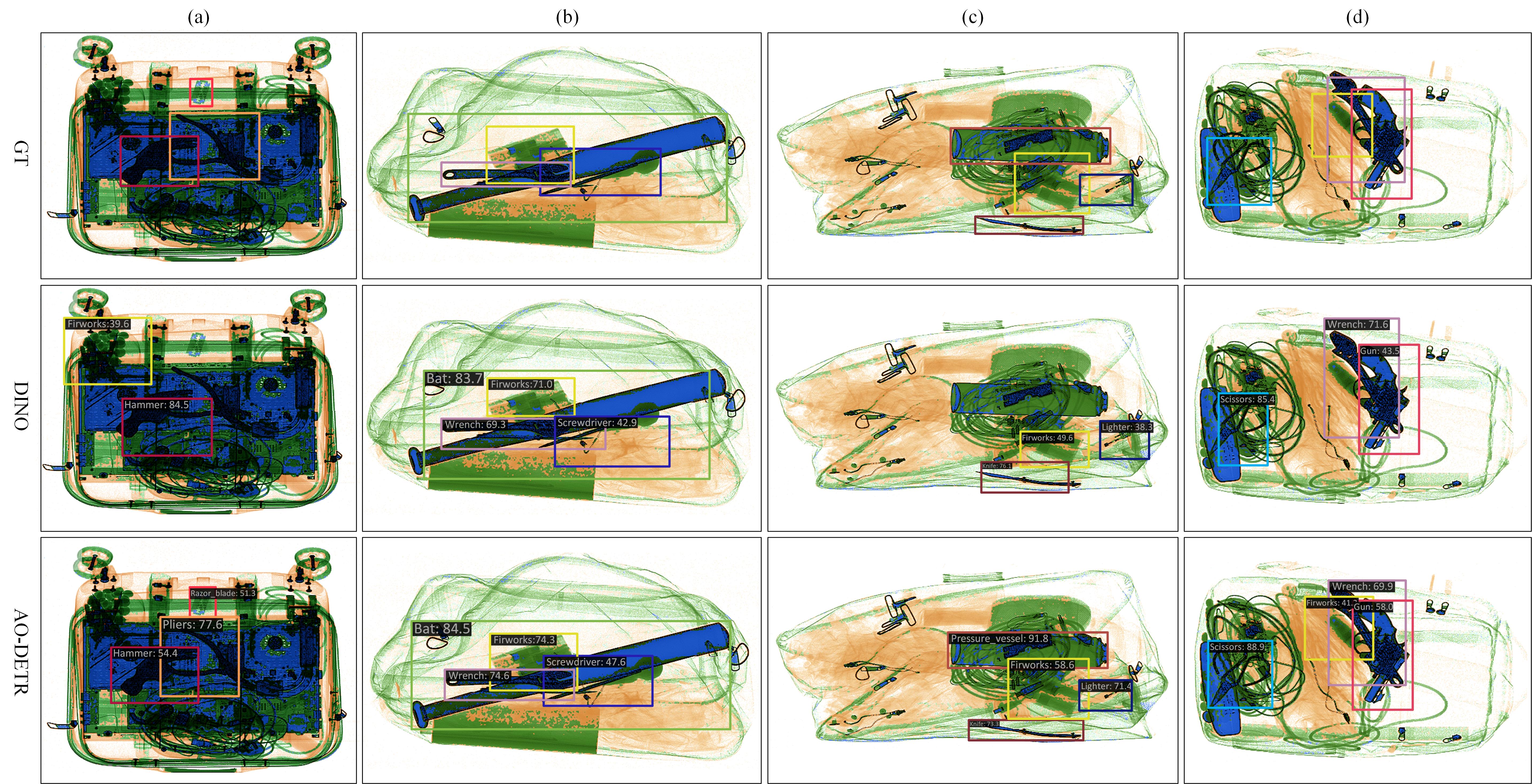}
    \caption{
    Comparison of X-ray prohibited item detection results on the PIXray~\cite{PIXray} dataset between the proposed AO-DETR and the DINO. 
    The row "GT" represents four typical X-ray prohibited item images with overlapping phenomenon, each with annotated ground truth boxes. The rows "DINO" and "AO-DETR" correspond to their respective detection results. To achieve optimal display effectiveness, we have standardized the color and category relationships between ground truth boxes and predicted boxes. For instance, yellow boxes denote fireworks, while bright red boxes signify razor blades.
    }
    \label{figure detection results visualization}
\end{figure*}
\subsection{Visualization Analysis}
\subsubsection{Visualization Analysis of Sampling Points}

As shown in Fig.~\ref{figure sampling points and reference points}, we take three X-ray images containing different categories of prohibited items as examples. From the last layer of the decoder, we select one category-specific object query from each of the four category-specific object query groups responsible for batteries, pressure vessels, fireworks, and razor blades. Then, we visualize their corresponding sampling points, reference points, and localization results.
Overall, the category-specific object query prioritizes perceiving and focusing on regions in the images that exhibit the highest similarity to the features of prohibited items of its responsible category. Taking the category-specific object query for batteries as an example, it recognizes batteries in row (a) and row (c), while in row (b), it attends to the top of a pressure vessel, which bears the highest similarity to a battery. Thanks to category-specific one-to-one matching, even though the category-specific object query for batteries observes the top of a pressure vessel in row (b), it can still discern that the features it attends to do not belong to a battery, leading to the decision to withhold output predictions.
Furthermore, the stability of category-specific object queries for category matching is remarkably high. In Fig.~\ref{figure sampling points and reference points}, there is no occurrence of a category-specific object query for category A focusing on and predicting prohibited items of category B. In row (a) and row (c), fireworks overlap significantly with background features, yet they are still accurately covered by the sampling points of the corresponding category-specific query. Moreover, when looking at the overall distribution of sampling points, those for batteries and razor blades are consistently densely concentrated in small areas, while those for fireworks and pressure vessels are consistently sparsely distributed in larger regions. This suggests that category-specific object queries extract features based on the size and shape characteristics of their responsible categories during the prediction process. In conclusion, we have successfully clarified the category semantics of object queries, using this as an opportunity to assist the network in identifying target objects in overlapping foreground-background scenarios.
\subsubsection{Visualization Analysis of Detection Results}
In Fig.~\ref{figure detection results visualization}, we present a comparative analysis of the detection results between DINO with ResNet-50 and AO-DETR with ResNet-50, using four X-ray images containing different prohibited items as examples.
We enumerate two adverse effects caused by overlapping phenomena. One is the feature coupling resulting from the overlap of prohibited items and the background, leading to missed detections. For instance, in column (a), a pair of pliers is missed by DINO due to the overlap. The other effect is the edge blurring caused by overlap, subsequently leading to inaccurate edge localization. In column (b), DINO inaccurately locates the overlapping part of fireworks and a screwdriver with a baseball bat. Column (c) and column (d) depict scenarios where both missed detections and inaccurate localization occur simultaneously.
In comparison, AO-DETR demonstrates accurate localization of prohibited items that heavily overlap with the background, and achieves higher recall rates, which showcases outstanding performance.

\section{Conclusions}\label{Conclusions}
In this paper, we first conduct an in-depth analysis of the two major challenges in the field of X-ray image prohibited item detection. Subsequently, we explore how to enhance general object detectors based on the characteristics of X-ray images. Overall, we improve the state-of-the-art DETR-like model in the general object detection domain, DINO, and introduce the AO-DETR series models.
Specifically, we propose the CSA strategy to enhance the anti-overlapping feature extraction capability for specific category foregrounds by constraining the object classes assigned to category-specific queries during the training phase. Furthermore, by employing the proposed LFD scheme, we enhance the accuracy of reference boxes predicted by mid-level and high-level decoder layers through dense gradient transmission, ultimately improving the ability to perceive blurry edges of models. Extensive experiments on the PIXray and OPIXray datasets demonstrate that our two novel methods can significantly enhance detection performance. Additionally, our AO-DETR series models outperform state-of-the-art detectors for various requirements.

\bibliographystyle{IEEEtran}











\bibliography{References}

 




\begin{IEEEbiography}[{\includegraphics[width=1in,height=1.25in,clip,keepaspectratio]{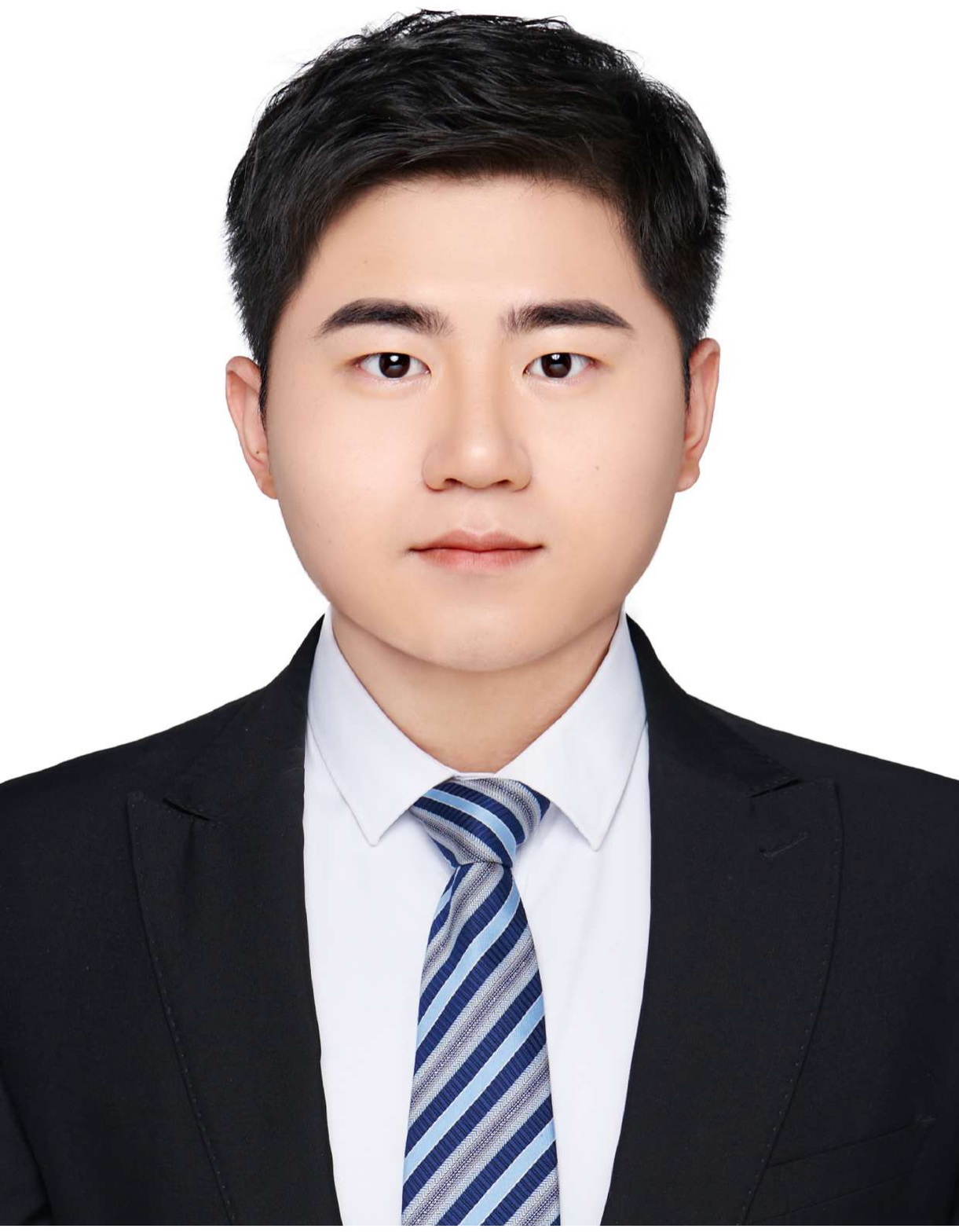}}]{Mingyuan Li}
received the B.E. degree from Northeastern University, China, in 2021. He is currently pursuing the Ph.D. degree with the College of Information Science and Engineering, Northeastern University, Shenyang, China. 

His research interests include deep learning, object detection, and prohibited item detection in X-ray images.
\end{IEEEbiography}

\begin{IEEEbiography}[{\includegraphics[width=1in,height=1.25in,clip,keepaspectratio]{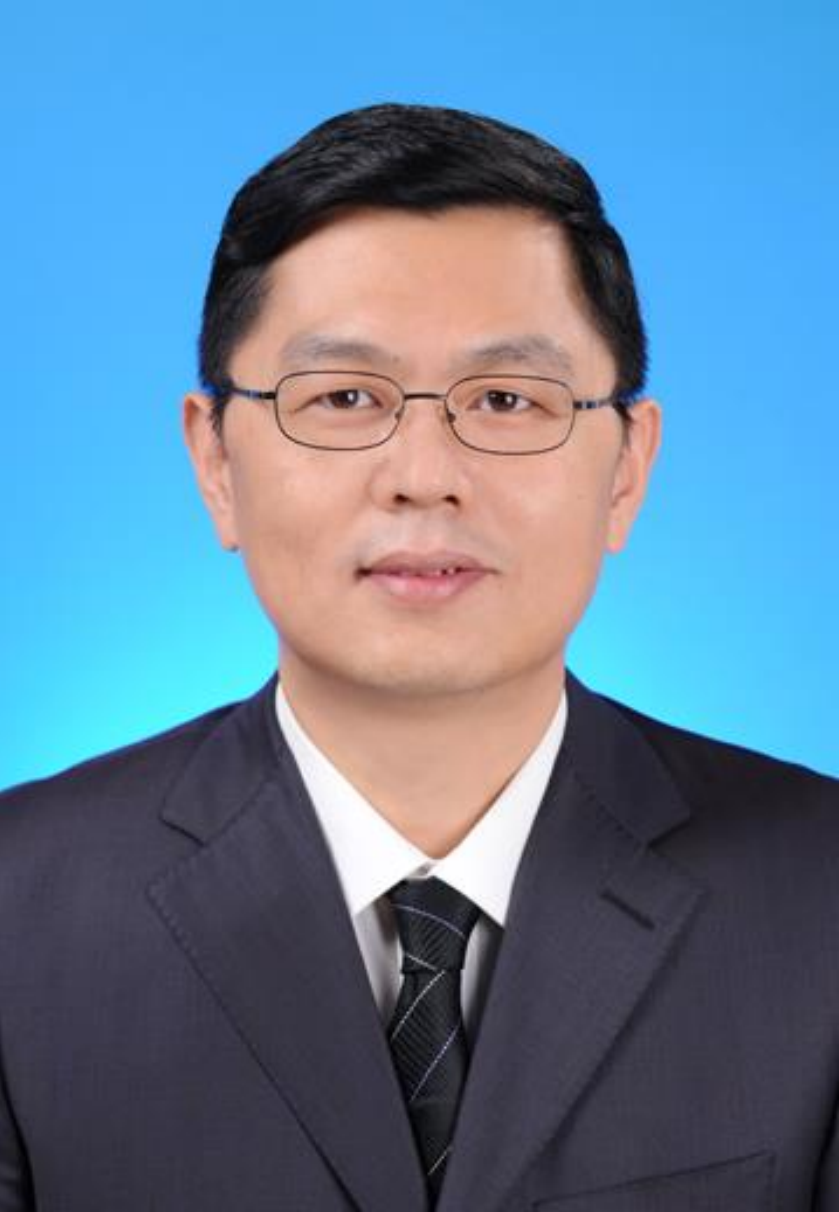}}]{Tong Jia} received the B.E. degree in computer science and the Ph.D. degree in pattern identification and intelligent system from Northeastern University, China, in 1998 and 2008, respectively. From 2012 to
2013, he was an International Visiting Scholar with the Department of Electronic Engineering, Michigan State University, USA.

He is currently a Professor with the College of Information Science and Engineering, Northeastern University, Shenyang, China. His research interests include computer/machine vision, image processing, and pattern identification.
\end{IEEEbiography}

\begin{IEEEbiography}[{\includegraphics[width=1in,height=1.25in,clip,keepaspectratio]{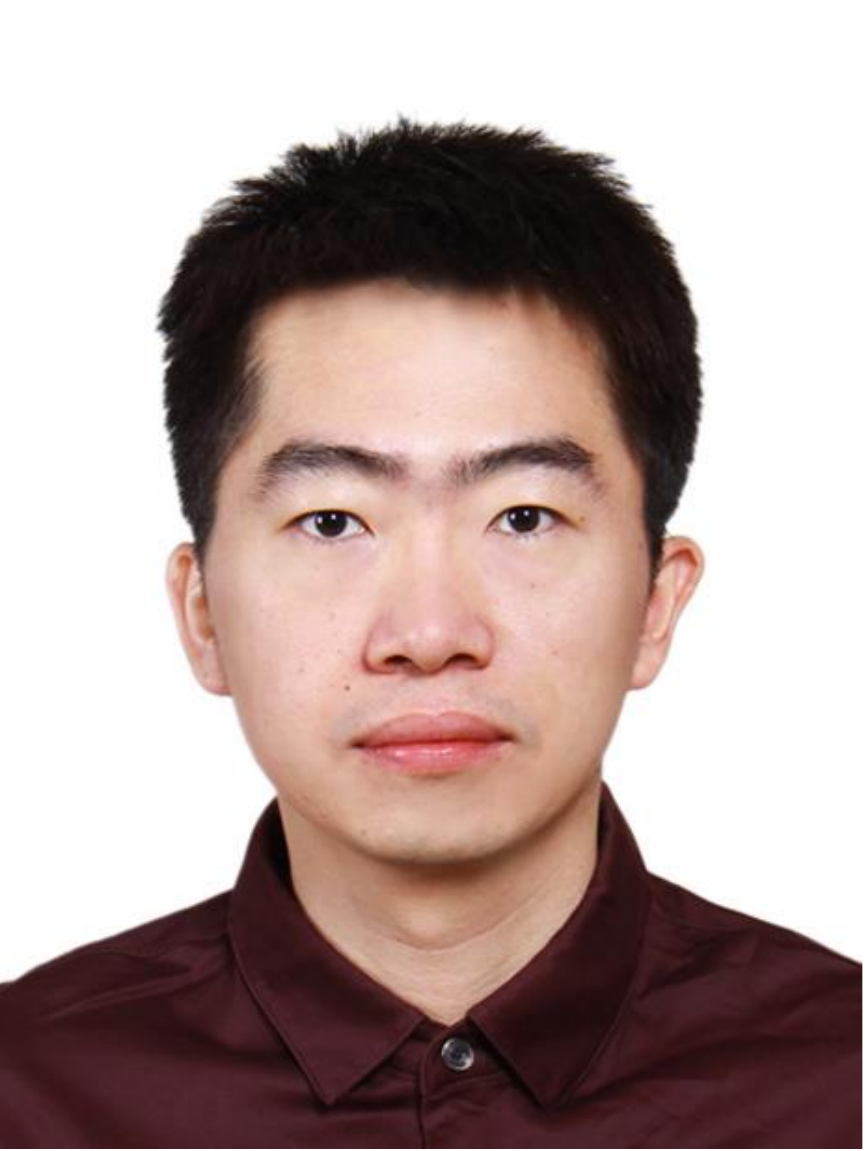}}]{Hao Wang}received the Ph.D. degree in the School
of Computer Science and Technology, Harbin Institute of Technology, Harbin, China, in 2022. He currently joins the Northeastern University, Shenyang,
China as a Lecturer with the College of Information
Science and Engineering.

His research interests include object detection, object segmentation, and related problems.
\end{IEEEbiography}

\begin{IEEEbiography}[{\includegraphics[width=1in,height=1.25in,clip,keepaspectratio]{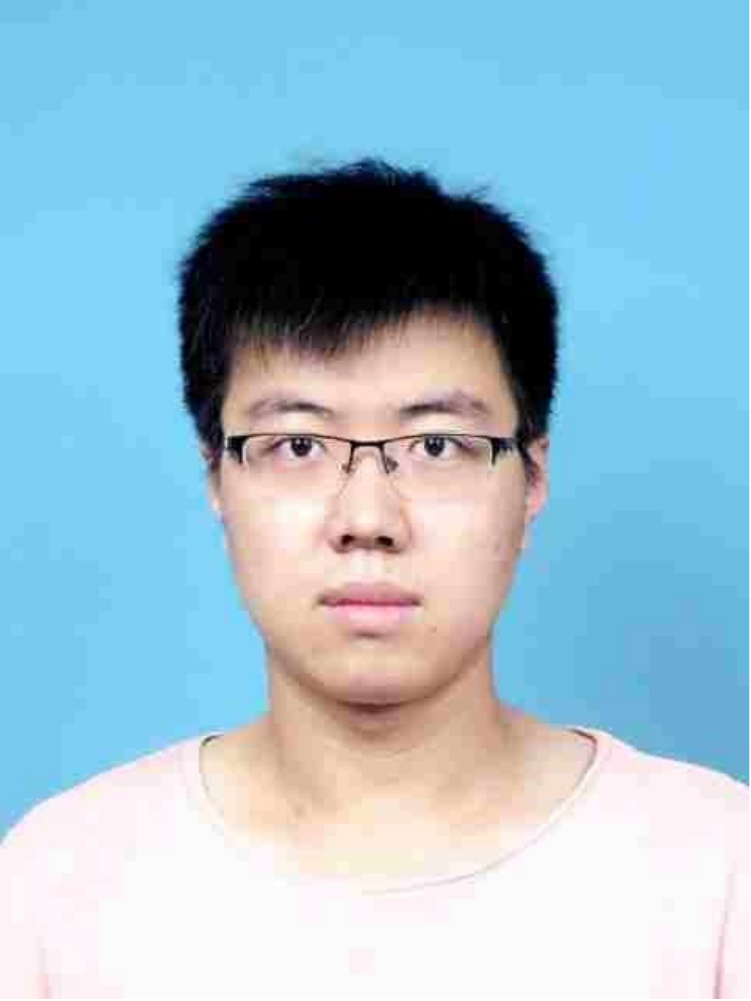}}]{Bowen Ma}
received the B.E. and M.S. degrees from Northeastern University, China, in 2017 and 2020, respectively. He is currently pursuing the Ph.D. degree with the College of Information Science and Engineering, Northeastern University, Shenyang, China. 

His research interests include computer vision, deep learning, and image processing for X-ray testing.
\end{IEEEbiography}

\begin{IEEEbiography}[{\includegraphics[width=1in,height=1.25in,clip,keepaspectratio]{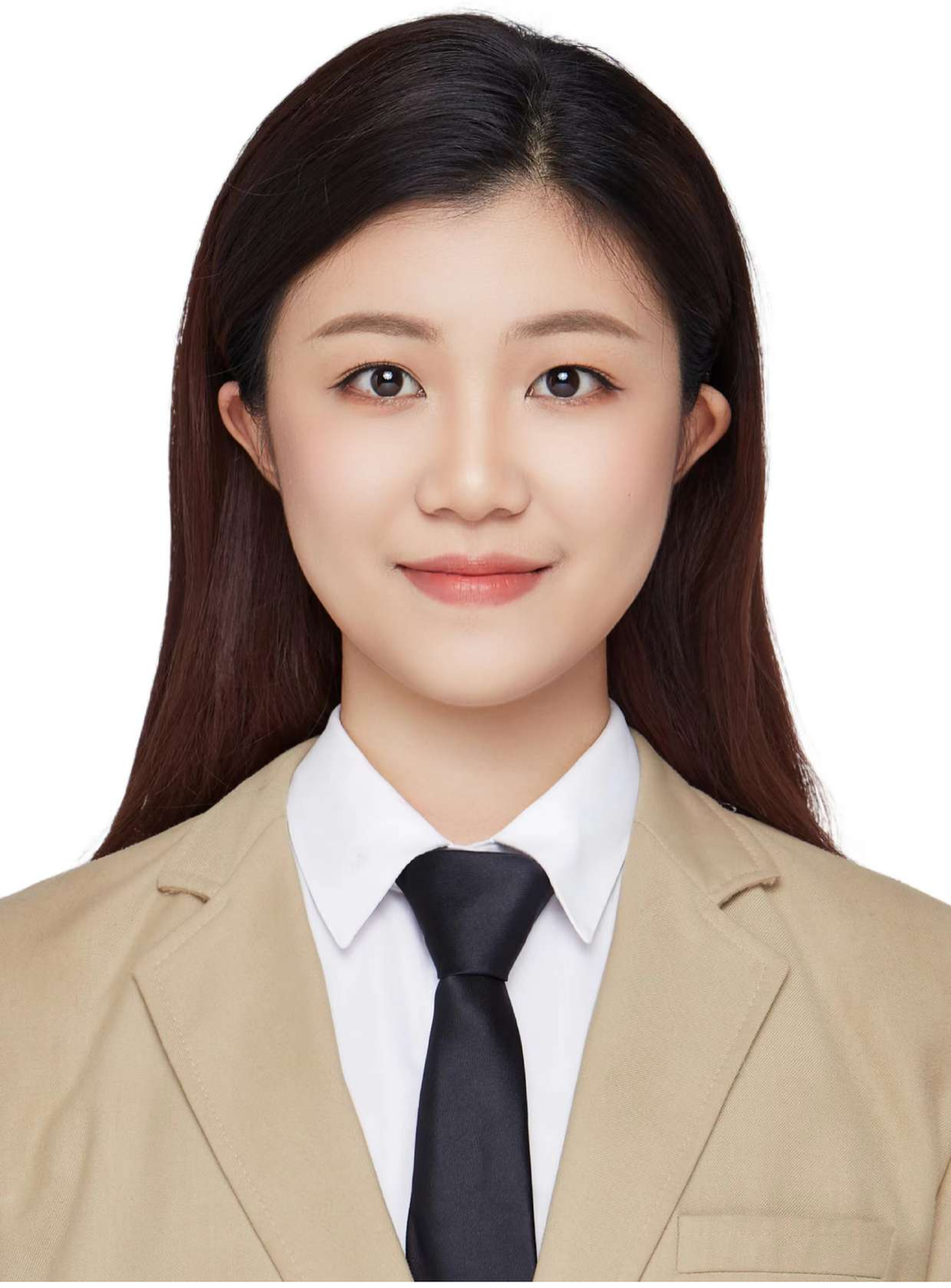}}]{Shuyang Lin} received the B.E. degree from Northeastern University, China, in 2022. She is currently pursuing the Ph.D. degree with the College of Information Science and Engineering, Northeastern University, Shenyang, China.
Her research interests include deep learning, X-ray prohibited item detection, open-vocabulary object detection and related problems.
\end{IEEEbiography}

\begin{IEEEbiography}[{\includegraphics[width=1in,height=1.25in,clip,keepaspectratio]{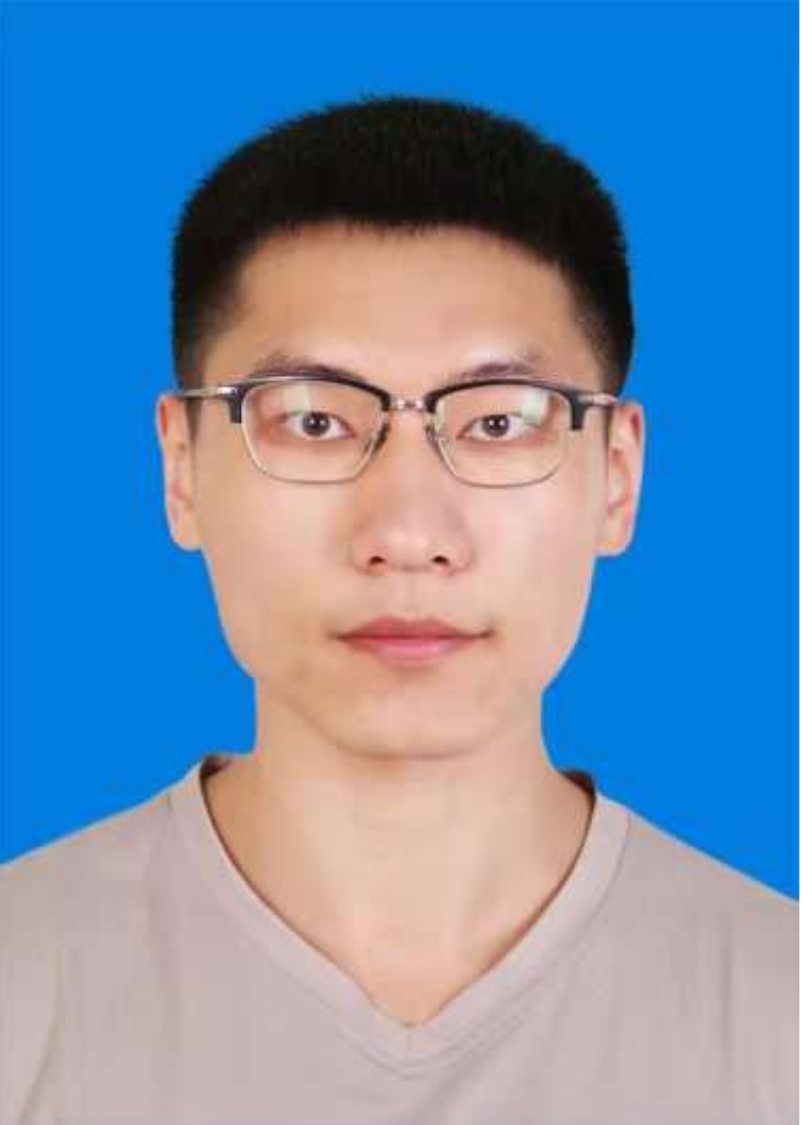}}]{Da Cai}
received the M.S. degrees
from Northeastern University, China, in 2021. He is currently pursuing
the Ph.D. degree with the College of Information
Science and Engineering, Northeastern University, Shenyang, China

His research interests include computer vision, deep learning and image processing for X-ray testing.
\end{IEEEbiography}

\begin{IEEEbiography}[{\includegraphics[width=1in,height=1.25in,clip,keepaspectratio]{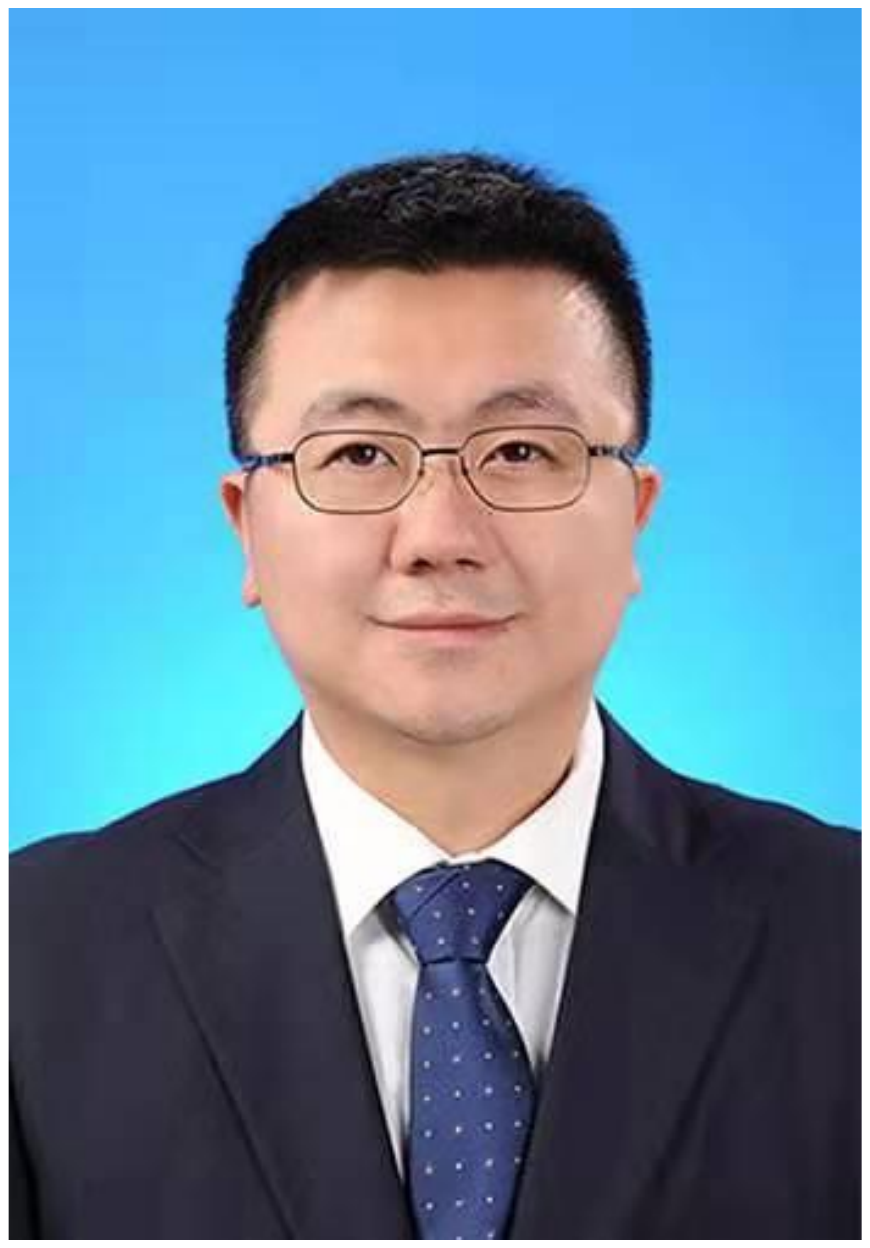}}]{Dongyue Chen} received the B.S. and Ph.D. degrees from the Department of Electronic Engineering, Fudan University, China, in 2002 and 2007, respectively. From 2014 to 2015, he was an International Visiting Scholar with the Perelman School of Medicine, University of Pennsylvania, Philadelphia, PA, USA.

He is currently a Professor with the College of Information Science and Engineering, Northeastern University, Shenyang, China. His research interests include biologically motivated visual modeling, computer vision, and image processing.
\end{IEEEbiography}

\vfill

\end{document}